\documentclass[a4wide]{article}

\usepackage{textcomp}
\usepackage[utf8]{inputenc}
\usepackage[T1]{fontenc}

\usepackage[round]{natbib}

\usepackage{amssymb}
\usepackage{amsmath}
\usepackage{amsthm}
\usepackage{mathtools}
\usepackage{algorithm,algorithmic}
\usepackage{stmaryrd}
\usepackage{booktabs}
\usepackage{color}
\usepackage{nicefrac}
\usepackage{multirow}
\usepackage{lipsum}

\DeclareMathOperator*{\argmin}{\mathrm{arg\,min}}
\DeclareMathOperator*{\argmax}{\mathrm{arg\,max}}

\newtheorem{theorem}{Theorem}[section]

\newtheorem{definition}[theorem]{Definition}

\usepackage[breaklinks,colorlinks]{hyperref}
\usepackage[dvipsnames]{xcolor}
\hypersetup{
    colorlinks,
    linkcolor={red!60!black},
    citecolor={green!40!black}
}

\usepackage[capitalize]{cleveref}
\newcommand{\q}[1]{``#1''}
\crefname{section}{Sec.}{Secs.}
\Crefname{section}{Section}{Sections}
\Crefname{table}{Table}{Tables}
\crefname{table}{Tab.}{Tabs.}

\usepackage{natbib}

\usepackage{authblk}

\usepackage{bbm}
\usepackage{tikz}
\usepackage{pgfplots}
\pgfplotsset{compat=1.16}
\usepgfplotslibrary{groupplots}
\usepgfplotslibrary{fillbetween}
\usepackage{adjustbox}
\usepackage{wrapfig}
\usepackage{setspace}
\usepackage[font=small]{caption}

\usepackage{url}
\usepackage{amsfonts}
\usepackage[nopatch=footnote]{microtype}
\usepackage{xcolor} 
\usepackage{soul}

\definecolor{plotcolor1}{HTML}{e41a1c}
\definecolor{plotcolor2}{HTML}{377eb8}
\definecolor{plotcolor3}{HTML}{de8b5d}
\definecolor{plotcolor4}{HTML}{e7298a}
\definecolor{plotcolor5}{HTML}{ff7f00}
\definecolor{plotcolor6}{HTML}{666655}
\definecolor{plotcolor7}{HTML}{5679b4}

\usetikzlibrary{arrows}
\usetikzlibrary{shapes.misc}

\tikzset{cross/.style={cross out, draw=black, minimum size=2*(#1-\pgflinewidth), inner sep=0pt, outer sep=0pt},
cross/.default={1pt}}

\usepackage{graphicx}
\usepackage{gensymb}
\usepackage{caption}
\usepackage{subcaption}
\usepackage{pgfplotstable}

\usepackage{microtype}

\title{Training on Plausible Counterfactuals Removes \\ Spurious Correlations}
\vspace{7mm}
\author[1,2]{Shpresim Sadiku}
\author[1,2]{Kartikeya Chitranshi}
\author[2,3]{Hiroshi Kera}
\author[1,2]{Sebastian Pokutta}
\affil[1]{\small Technische Universität Berlin, Institute of Mathematics}
\affil[2]{Zuse Institute Berlin, Department AIS2T,
\emph{lastname}@zib.de}
\affil[3]{Chiba University, Institute for Advanced Academic Research, kera@chiba-u.jp}
\date{}
\begin{document}
\maketitle

\vspace{5mm}

\begin{center}
\begin{minipage}{0.85\textwidth}
\begin{center}
 \textbf{Abstract}
\end{center}
 {\small Plausible counterfactual explanations (p-CFEs) are perturbations that minimally modify inputs to change classifier decisions while remaining plausible under the data distribution. In this study, we demonstrate that classifiers can be trained on p-CFEs labeled with induced \emph{incorrect} target classes to classify unperturbed inputs with the original labels. While previous studies have shown that such learning is possible with adversarial perturbations, we extend this paradigm to p-CFEs.
Interestingly, our experiments reveal that learning from p-CFEs is even more effective: the resulting classifiers achieve not only high in-distribution accuracy but also exhibit significantly reduced bias with respect to spurious correlations.}
\end{minipage}
\end{center}

\vspace{0mm}

\section{Introduction}
\label{sec:introduction}

Altering a classifier’s prediction through minimal input perturbations has yielded valuable insights into the decision-making processes of machine learning models. \emph{Adversarial attacks}~\citep{szegedy2013intriguing}, for instance, have demonstrated the unexpected vulnerability of well-trained models to imperceptibly small perturbations, and various forms of such perturbations have been found through extensive studies~\citep{wachter2017counterfactual, madry2018towards}. In contrast, \emph{plausible counterfactual explanations} (p-CFEs) are minimal perturbations that alter classifications in a semantically coherent manner. Designed to align with the data manifold and to be interpretable, p-CFEs offer visual explanations of model predictions through \q{what-if} scenarios~\citep{zhang2023density, 2024_SadikuEtAl_Counterfactualexplanations}. Despite differing in their constraints, adversarial attacks and p-CFEs share a common objective: altering model predictions through minimal input changes-—suggesting potential synergies that remain underexplored in the literature.

Recently, \citet{kumano2024theoretical} revisited a seminal study by \citet{ilyas2019adversarial} and theoretically justified their observations: adversarial perturbations, although seemingly subtle and meaningless, actually contain generalizable, class-specific features---a model trained on them labeled with their \emph{induced} (incorrect) classes can successfully classify clean images into \emph{original} classes. 
Given the structural similarity between adversarial examples and p-CFEs as minimal input perturbations, it is natural to ask whether the representational richness observed in adversarial examples also applies to p-CFEs. However, the defining characteristic of p-CFEs---\emph{plausibility}---may lead to fundamentally different outcomes. 

This study addresses \emph{learning from p-CFEs} and empirically demonstrates that it is more effective than learning from adversarial perturbations, achieving higher in-distribution and out-of-distribution accuracy in the presence of \emph{spurious correlations}--features that correlate with the label during training but are semantically irrelevant.  Our results suggest that p-CFEs not only induce prediction flips but also guide models toward learning features that better reflect the true, semantic structure of the data.

\paragraph{Contributions}
\begin{enumerate}
    \item \textbf{New instance of learning from perturbations.} We demonstrate that the insightful observation of learning from adversarial perturbations by \citet{ilyas2019adversarial} generalizes to p-CFEs, highlighting the broader applicability of the \textit{learning from perturbations} paradigm.

    \item \textbf{High classification accuracy.} Our experiments show that learning from p-CFEs~\citep{2024_SadikuEtAl_Counterfactualexplanations} yields high classification accuracy on original samples—comparable to models trained on adversarial perturbations (including $\ell_{2}$ and $\ell_{\infty}$ PGD, and $\ell_{2}$ CFEs).

    \item \textbf{Removing spurious correlations.} The experiments further reveal that learning from p-CFEs significantly outperforms learning from perturbations in mitigating spurious correlations. On the WaterBirds dataset, where evaluation sets are designed with strong spurious correlations, learning from p-CFEs even surpasses standard (noise-free) training by 12\,\% in worst-group accuracy.
\end{enumerate}

\begin{figure*}
    \centering
    \setlength{\tabcolsep}{2pt} 
    \begin{tabular}{@{}c@{\hspace{2pt}}c@{\hspace{2pt}}c@{\hspace{2pt}}c@{\hspace{2pt}}c@{\hspace{2pt}}c@{}}
         $\mathcal{D}$ & $\tilde{\mathcal{D}}_{\textnormal{PGD\ } \ell_2}$ & $\tilde{\mathcal{D}}_{\textnormal{PGD\ } \ell_\infty}$ & $\tilde{\mathcal{D}}_{\textnormal{CFE\ } \ell_2}$ & $\tilde{\mathcal{D}}_{\textnormal{p-CFE\ } \ell_0}$ \\ 
        
         \includegraphics[width=0.14\textwidth]{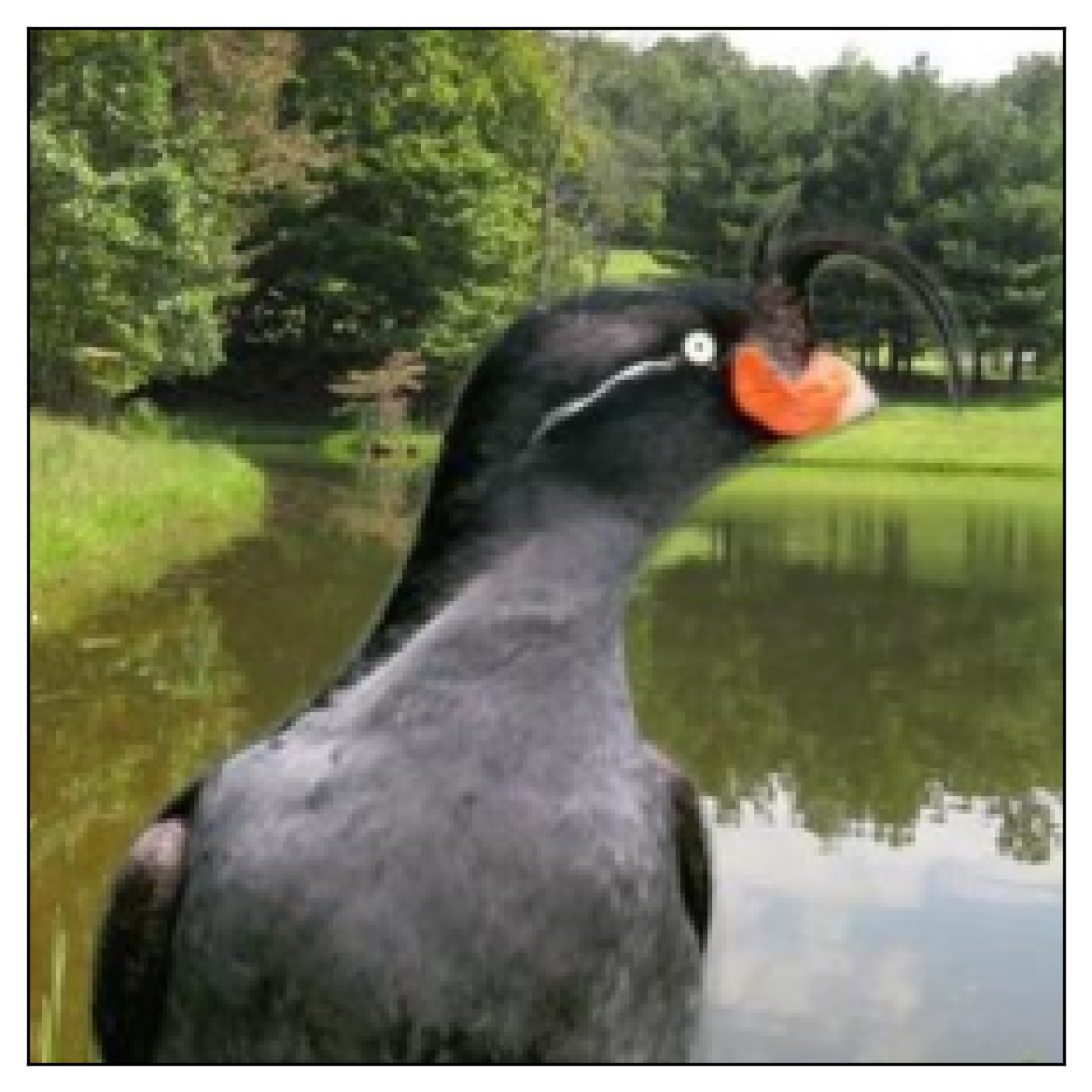} &
        \includegraphics[width=0.14\textwidth]{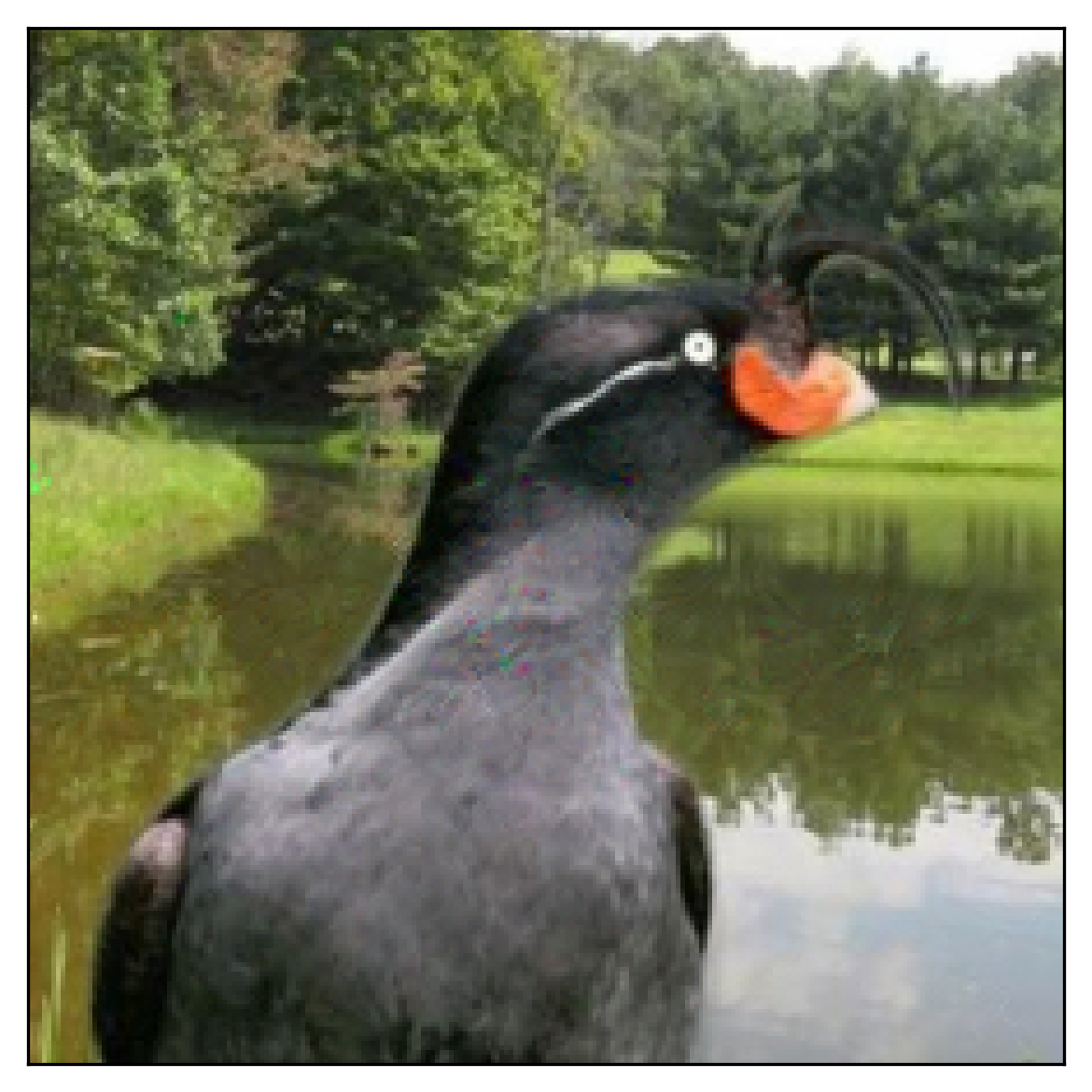} &
        \includegraphics[width=0.14\textwidth]{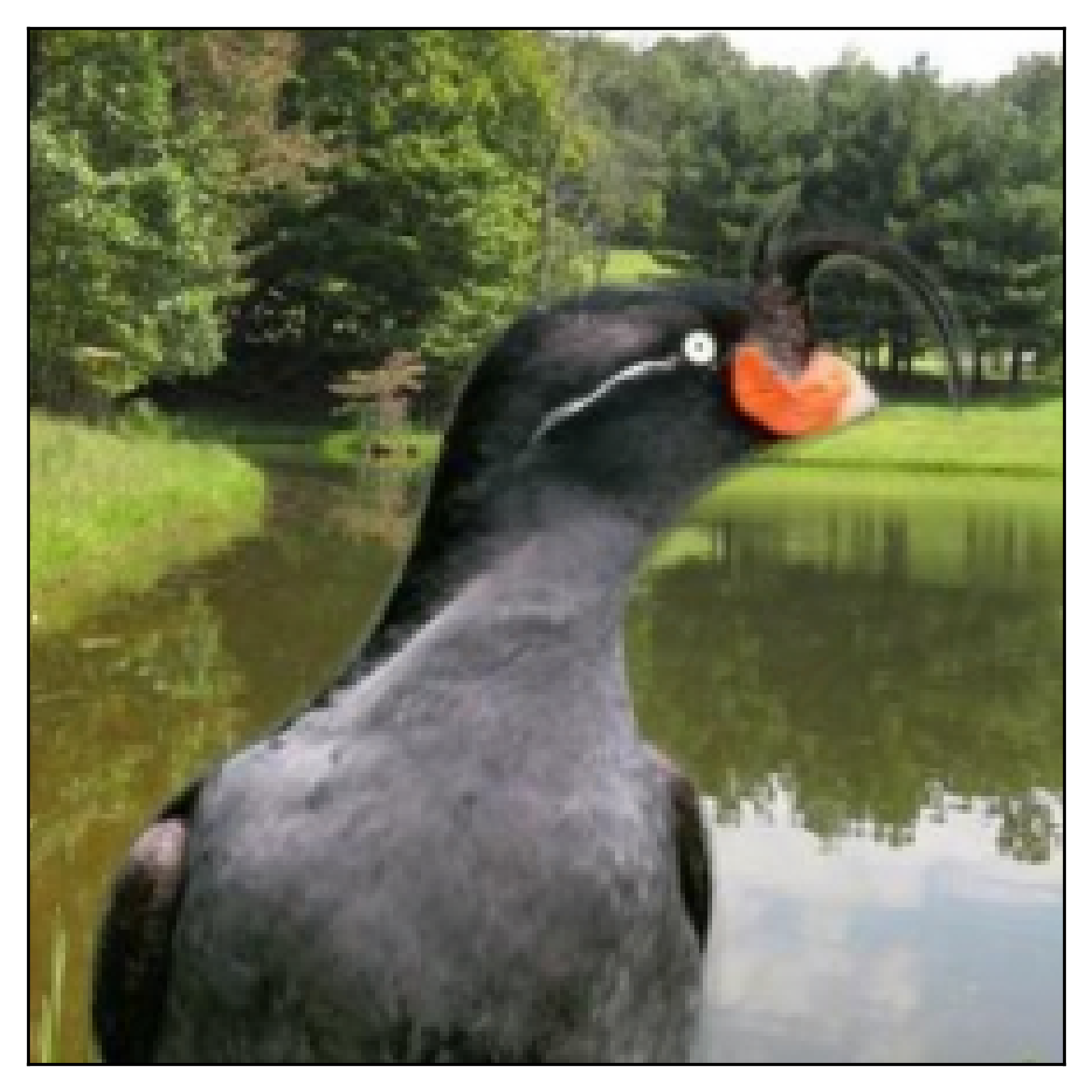} &
        \includegraphics[width=0.14\textwidth]{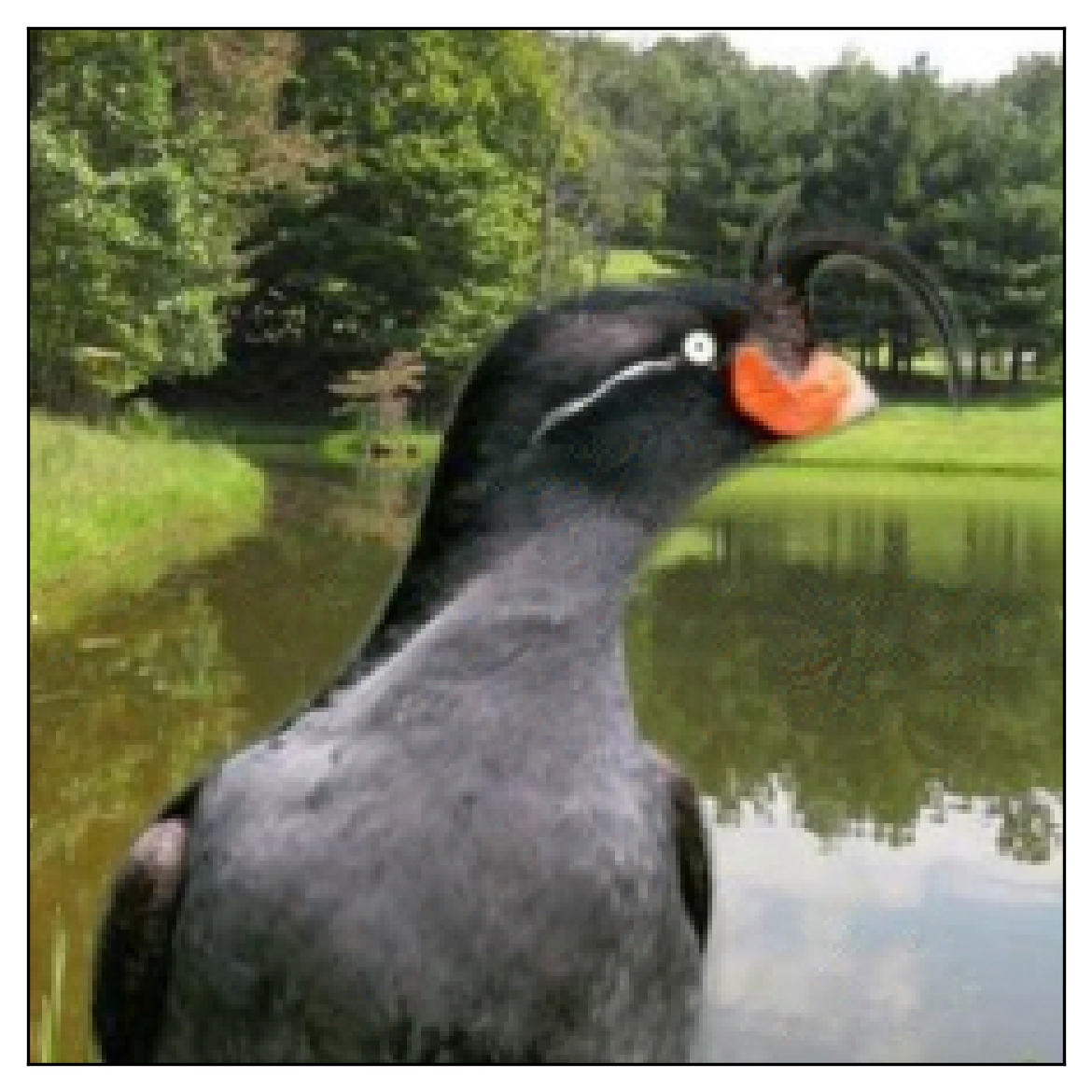} &
        \includegraphics[width=0.14\textwidth]{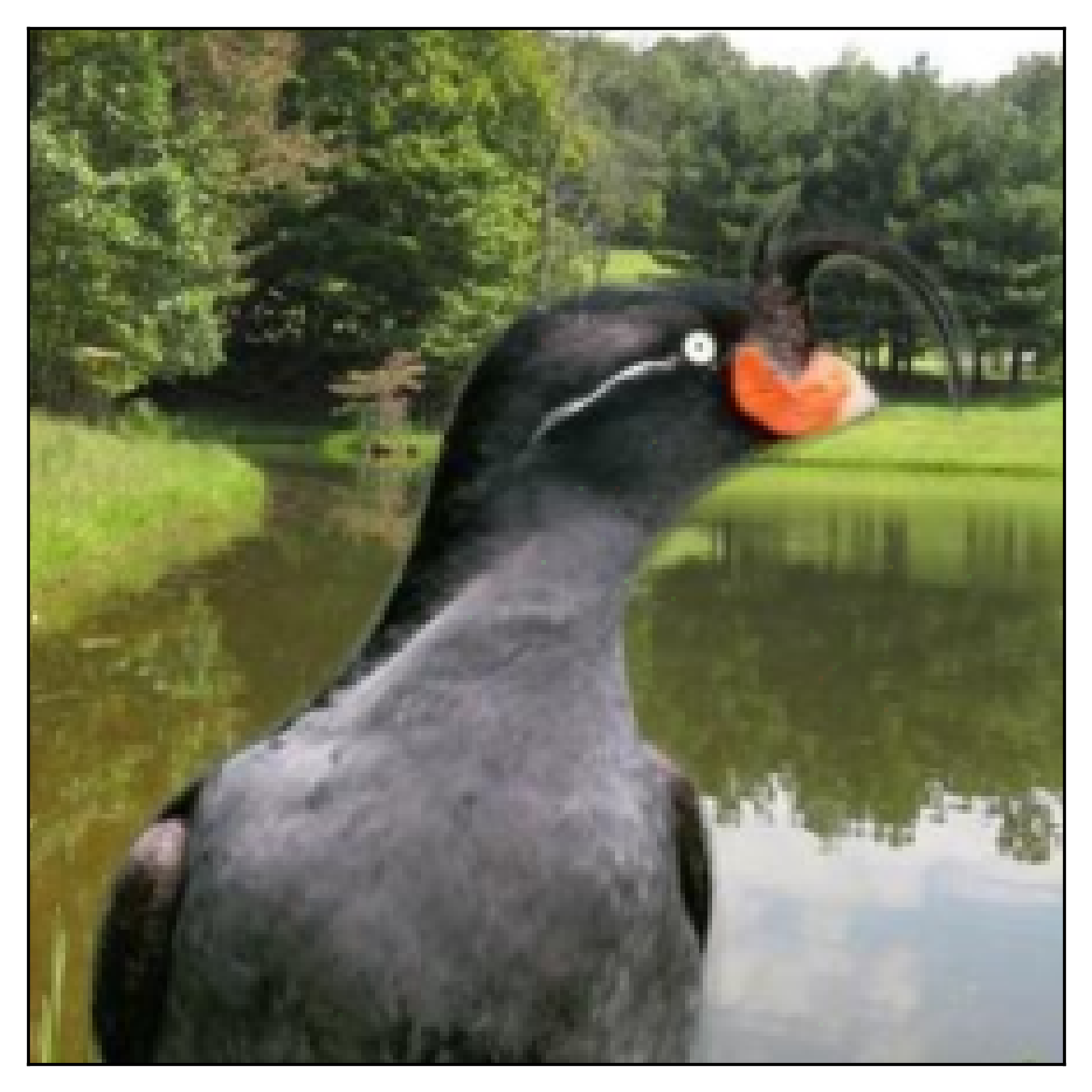} \\
        & 
        \includegraphics[width=0.135\textwidth]{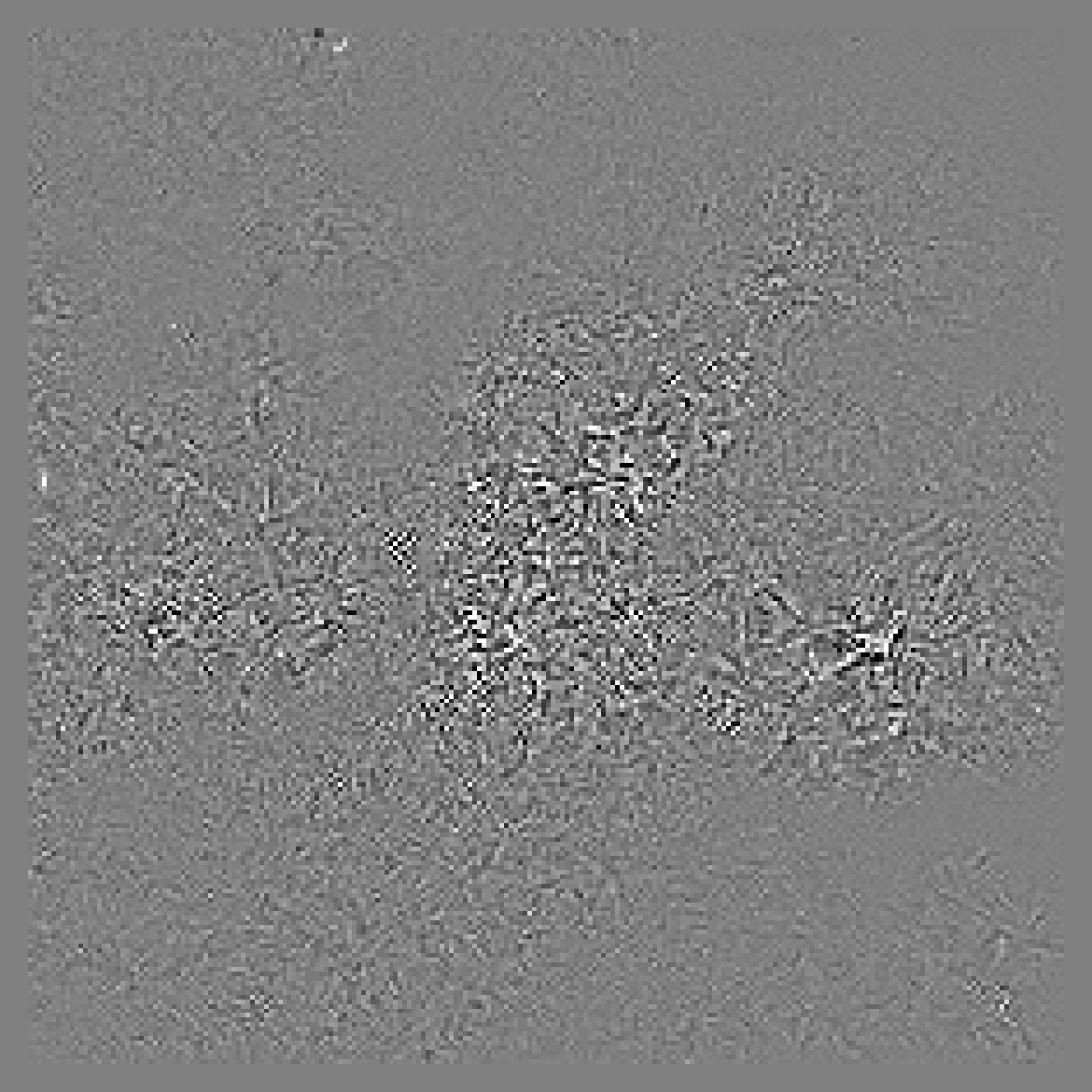} &
        \includegraphics[width=0.135\textwidth]{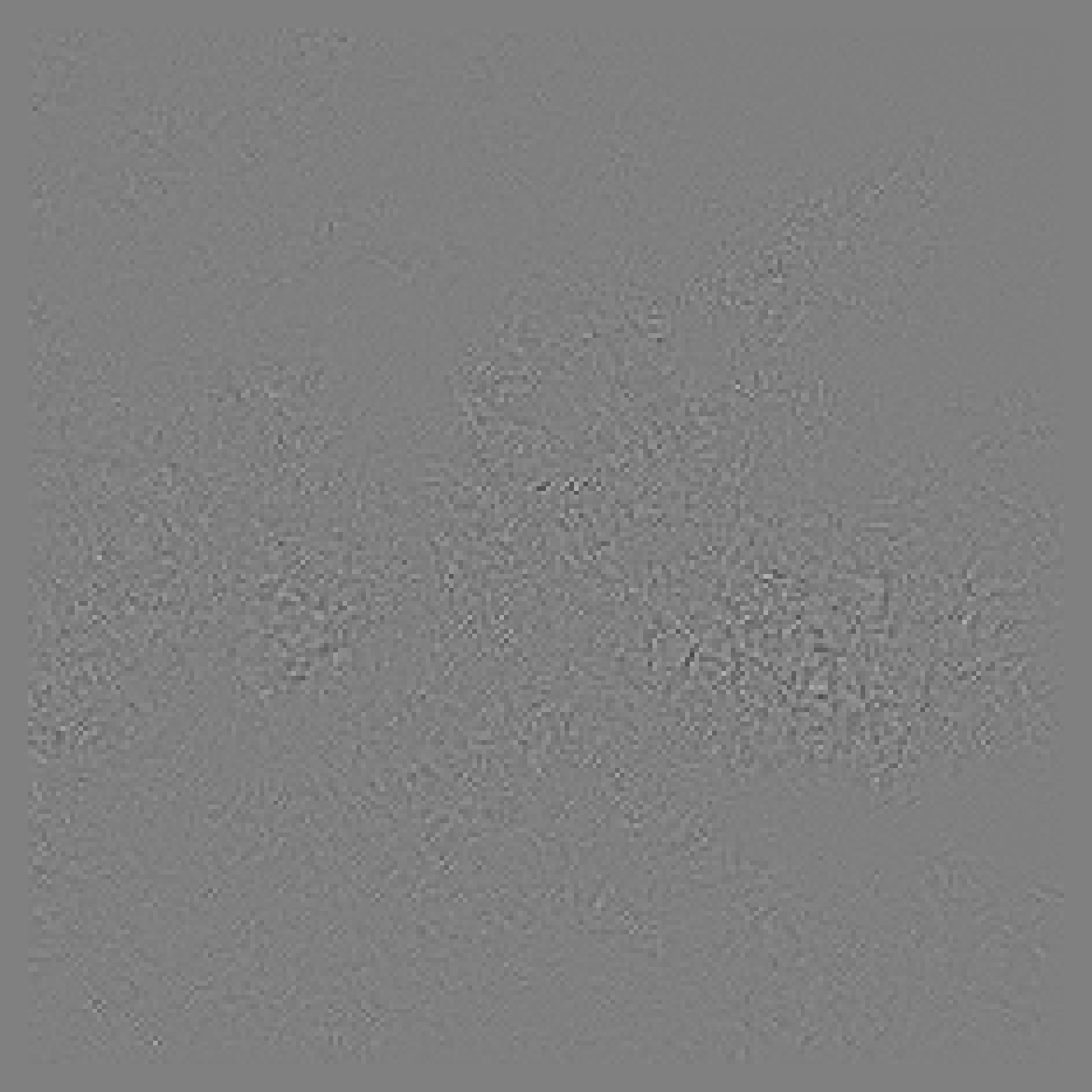} &
        \includegraphics[width=0.135\textwidth]{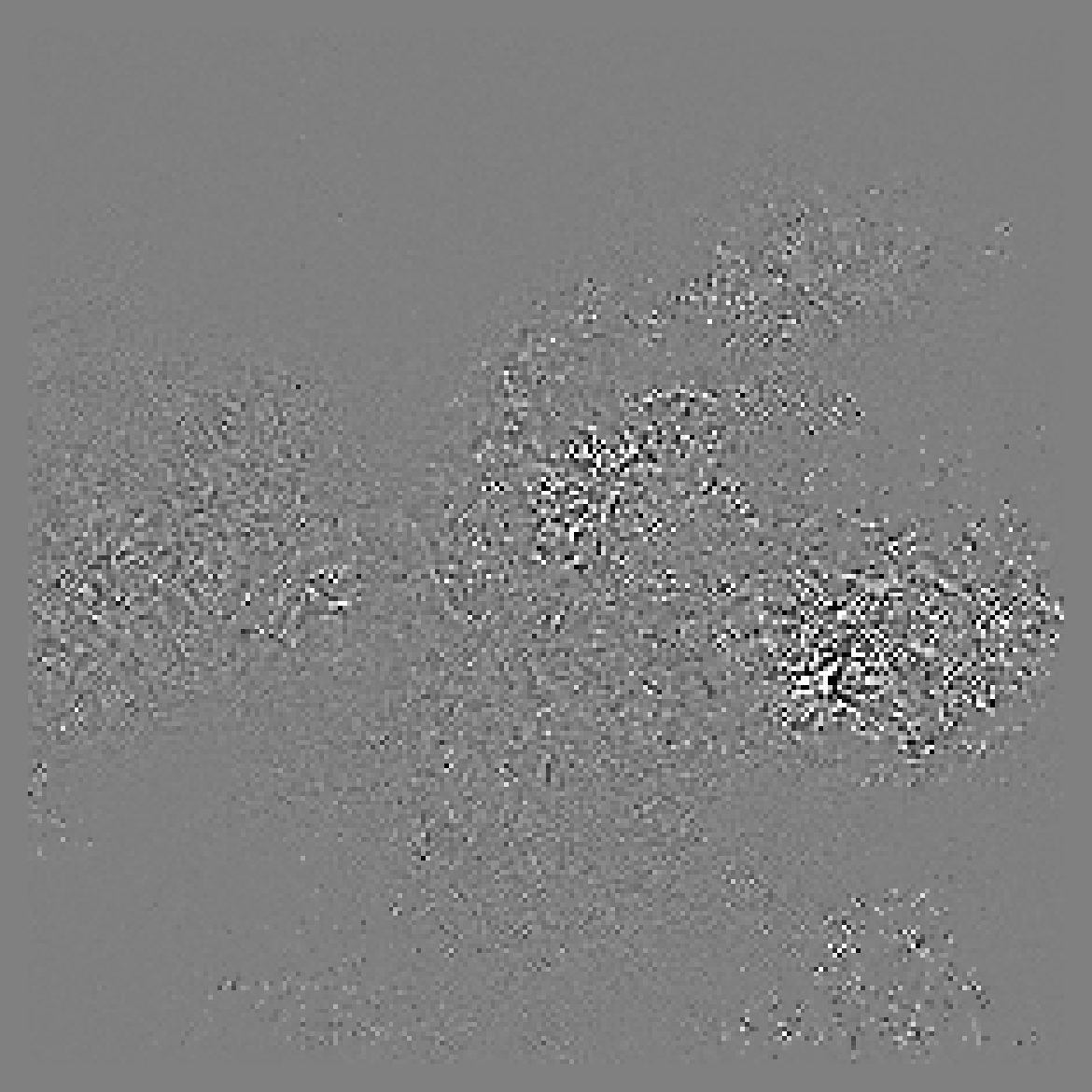} &
        \includegraphics[width=0.135\textwidth]{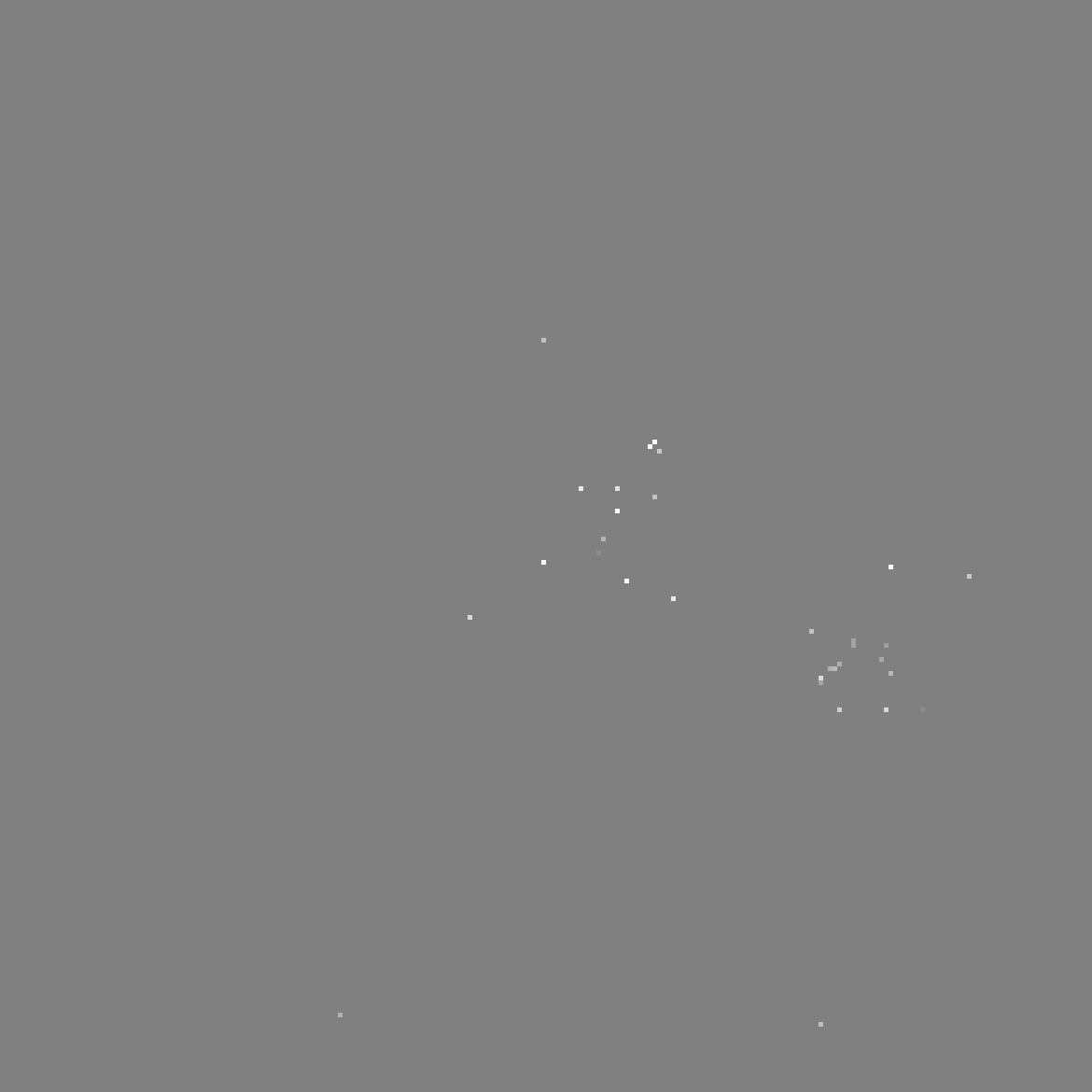} 
        
    \end{tabular}
    \caption{Random samples from our WaterBirds training set variants. The bottom row shows perturbations (magnified 40 times for visibility) applied to the original image on the left by different methods. The true label is \emph{water bird}; the target label is \emph{land bird}.}
    \label{trainingdata}
\end{figure*}

\section{Related Work}
\label{sec:relwork}
Small input perturbations have yielded various insightful observations that deepen our understanding of machine learning models. A prominent example is adversarial perturbations~\citep{szegedy2013intriguing}, which can easily fool seemingly strong classifiers with imperceptibly small changes, thereby questioning the reliability of machine learning models. Such perturbations exist in various forms~\citep{madry2018towards, xu2018structured, kazemi2023minimally}, and they are known to transfer across different models~\citep{xiaosen2023rethinking, 2023_SadikuWagnerPokutta_Groupwisesparseattacks}.
Counterfactual explanations (CFEs;~\citep{wachter2017counterfactual}) represent another line of research on small input perturbations. While early CFEs resembled adversarial perturbations, recent approaches increasingly emphasize alignment with the data manifold. In particular, \citet{2024_SadikuEtAl_Counterfactualexplanations} proposed a method for generating plausible counterfactuals (p-CFEs) via proximal gradient optimization, yielding perturbations that contain semantics aligned with the target class. However, such counterfactuals have primarily been used for interpretability purposes rather than as training data. 
%
Although a few studies have analyzed the connection between CFEs and adversarial perturbations~\citep{pawelczyk2022exploring, freiesleben2022intriguing}, the interaction between these two research directions remains limited. 

This study investigates whether observations in the literature of adversarial perturbations,  particularly learning from adversarial perturbations~\citep{ilyas2019adversarial,kumano2024theoretical,kumano2024wide}, extend to p-CFEs, and how the plausibility of p-CFEs gives rise to distinct outcomes.

\section{Preliminaries}
 Assume a binary classification setting where $\mathcal{X} \subseteq \mathbb{R}^d$ denotes the input space, $\mathcal{Y} = \{ \pm 1 \}$ denotes the set of possible class labels, and $\mathcal{D} = \{ (\boldsymbol{x}_i,y_i) \in \mathcal{X \times Y} \}_{i=1}^n$ is a training dataset consisting of $n$ independent and identically distributed data points generated from a joint density $\psi: \mathcal{X \times Y} \mapsto \mathbb{R}_+$. Furthermore, we define $q(\boldsymbol{x},y):=\psi(\boldsymbol{x}|y)$, which is the corresponding density of the inputs conditioned on the given label $y$.

 We let $f_{\theta}: \mathcal{X} \to \mathbb{R}^2$ denote a neural network classifier that takes a $d$-dimensional sample as input and outputs logits of the two classes. The final decision is denoted by $f(\boldsymbol{x}):=\argmax_i [f_{\theta}(\boldsymbol{x})]_i$. For $d \in \mathbb{N}$ let $[d] = \{ 1,...,d\}$. 

 \textbf{Standard Training.}
 With the exponential loss $\mathcal{L}(\boldsymbol{x},y):= \exp (-y \cdot f(\boldsymbol{x}))$ or logistic loss $\mathcal{L}(\boldsymbol{x},y):= \ln(1+\exp(-y \cdot f(\boldsymbol{x})))$, training a classifier is performed by minimizing a loss function $\hat{R} (\theta) := \sum_{i=1}^n \mathcal{L} (\boldsymbol{x}_i,y_i)/n$ from the training set via \emph{empirical risk minimization} (ERM).

\section{Learning from p-CFEs}

We now introduce a formal definition of learning from perturbations. 
 \begin{definition}[Learning from perturbations]
     Let $\mathcal{D} := \{(\boldsymbol{x}_i, y_i)\}_{i=1}^n$ be a training dataset, where each $\boldsymbol{x}_i$ is an input and $y_i$ is its corresponding label. Let $f$ be a classifier trained on $\mathcal{D}$ via standard training. For each $i$, a perturbed example $\tilde{\boldsymbol{x}}_i$ is generated to increase the probability of a target label $\tilde{y}_i \neq y_i$ under $f$, resulting in a perturbed dataset $\tilde{\mathcal{D}} := \{ (\tilde{\boldsymbol{x}}_i, \tilde{y}_i) \}_{i=1}^n$. Training a classifier from scratch on $\tilde{\mathcal{D}}$, is referred to as \emph{learning from perturbations}.
    \label{learningfromperturbations}
\end{definition}

Prior studies~\citep{ilyas2019adversarial,kumano2024theoretical,kumano2024wide} assume the perturbations to be adversarial ones, typically generated using targeted Projected Gradient Descent~(PGD;~\citep{madry2017towards}) via the following optimization for input $\boldsymbol{x}$, target label $\tilde{y}$, and perturbation budget $\epsilon > 0$

\begin{align}
    \min_{\tilde{\boldsymbol{x}} \in \mathcal{X}} \mathcal{L}(\tilde{\boldsymbol{x}}, \tilde{y}) \quad \text{s.t.}\quad \| \tilde{\boldsymbol{x}} - \boldsymbol{x} \|_p \leq \epsilon,
\end{align}

where $\| \cdot \|_p$ denotes the $\ell_p$ norm. Crucially, this differs from standard \emph{adversarial training}: each training sample $\tilde{\boldsymbol{x}}_i$ is paired with a \textit{target incorrect label} $\tilde{y}_i$, while evaluation is performed on the clean input $\boldsymbol{x}_i$ with the \textit{original label} $y$.

\def\offs{1.9}
\def\imgwidth{0.165}
\begin{wrapfigure}[19]{r}{0.48\textwidth}
\vspace{0.35cm}
    \centering
    \adjustbox{max width=0.45\textwidth}{
    \begin{tikzpicture}

            \node at (\offs, \offs + 1.2) {\small Original};
            \node at (2.55*\offs, \offs + 1.235) {\small Grad-CAM};

            \node at (\offs, \offs) {\includegraphics[width=\imgwidth\textwidth, height=0.10\textwidth]{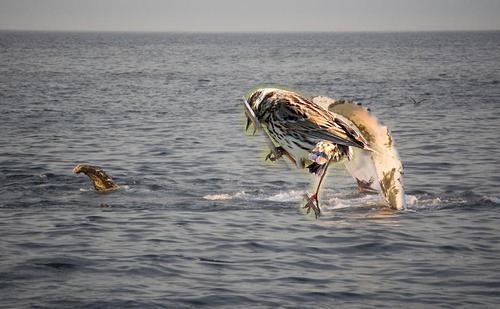}};
            \node at (2.55*\offs, \offs) {\includegraphics[width=\imgwidth\textwidth, height=0.10\textwidth]{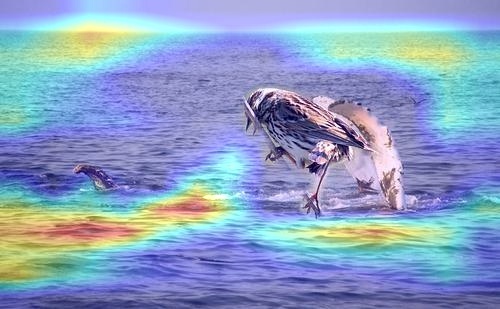}};

            \node at (\offs, 0.016*\offs) {\includegraphics[width=\imgwidth\textwidth, height=0.10\textwidth]{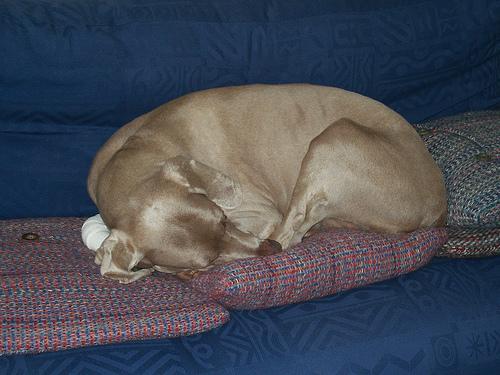}};
            \node at (2.55*\offs, 0.016*\offs) {\includegraphics[width=\imgwidth\textwidth, height=0.10\textwidth]{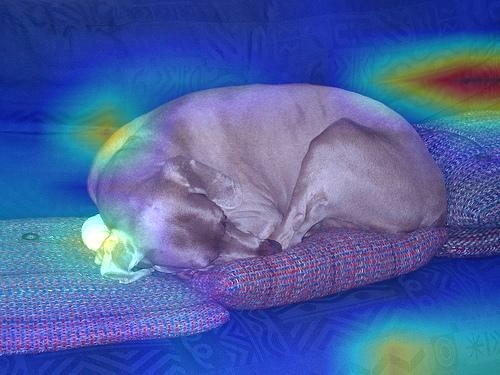}};
    \end{tikzpicture}
    }
    \caption{\textbf{Top row}: Original and Grad-CAM \citep{selvaraju2017grad} visualizations for a misclassified \emph{landbird} (with a water background) from the WaterBirds dataset—incorrectly predicted as a \emph{waterbird}. \textbf{Bottom row}: Original and Grad-CAM visualizations for a misclassified \emph{big dog} (with an indoor background) from the SpuCoAnimals dataset—incorrectly predicted as a \emph{small dog}.}
    \label{fig:waterbirds_Grad-CAM}
\end{wrapfigure}

We extend this idea to p-CFEs by perturbing each input-label pair $(\boldsymbol{x},y)$ using the method of \citet{2024_SadikuEtAl_Counterfactualexplanations}, formulated as the following unconstrained optimization problem

\begin{equation}
    \begin{aligned}
    \boldsymbol{\tilde{x}} :  =  \ \argmin_{\boldsymbol{x}^\prime \in \mathcal{A}}    & \ \|\boldsymbol{x}^\prime - \boldsymbol{x}\|_2^2  +\gamma \mathcal L(\boldsymbol{x}^\prime, \tilde{y}) \\ & - \tau \hat{q}(\boldsymbol{x}^\prime ,\tilde{y}) + \beta \| \boldsymbol{x}^\prime - \boldsymbol{x} \|_0,
    \end{aligned} \label{lagrangianclosestsparsedatacfe}
\end{equation}

where $\hat{q}(\,\cdot\,, \tilde{y})$ estimates the density of the target class $\tilde{y}$ in $\mathcal{X}$, and $\mathcal{A}:= \bigtimes_{i=1}^d [-\mathcal{A}_i, \mathcal{A}_i],$ with $\mathcal{A}_i \in \mathbb{R},$ defines the feature value range, either derived from the dataset or specified by the user. The parameters $\gamma, \tau, \beta > 0$ control tradeoffs for \emph{validity} (flipping the decision), \emph{plausibility} (staying on the data manifold), and \emph{sparsity} (minimizing feature changes), respectively.

\paragraph{Plausibility Term.}
The plausibility term $\hat{q}(\,\cdot\,, \tilde{y})$ only needs to be differentiable, enabling gradient-based optimization. For example, KDEs and GMMs are standard differentiable estimators~\citep{2024_SadikuEtAl_Counterfactualexplanations}. For higher-dimensional data, more expressive models like VAEs or GANs can be used \citep{van2021interpretable}. The method is compatible with any modern classifier that supports backpropagation, including large language models (e.g., GPT-$n$ with $n>2$) and vision models such as Swin Transformers with GELU activations \citep{liu2021swin}.

\paragraph{Adversarial vs. Counterfactual.}
The key distinction in \cref{lagrangianclosestsparsedatacfe} compared to adversarial attacks is the term minimizing the negative target-class density estimate, which encourages perturbed instances to lie on the target data manifold. By rewriting \cref{lagrangianclosestsparsedatacfe} as the sum of a smooth (possibly non-convex) term and a non-smooth term with a closed-form proximal operator, we adopt the solution strategy of \citet{2024_SadikuEtAl_Counterfactualexplanations}, using the accelerated proximal gradient method of \citet{beck2009fast} to generate p-CFEs.


\section{Experiments}

\paragraph{Datasets.}  We adopt two standard benchmark datasets that involve spurious correlations \citep{bender2023towards}.  The \textbf{WaterBirds}~\citep{sagawa2019distributionally} dataset has two labels, namely landbird and waterbird. The spurious correlation arises from the change in background (like a water bird on land or a landbird above or on water).
The \textbf{SpuCoAnimals}~\citep{joshi2023towards} dataset contains two categories: big dogs and small dogs. Spurious correlation arises on the assumption that big dogs were mostly outside and the small dogs inside the house. \cref{fig:waterbirds_Grad-CAM} visually illustrates spurious correlations in both datasets, with additional examples given in \cref{spuriousadditional}.
For a review on spurious correlations, see \citep{ye2024spurious}. 

\begin{figure*}[t]
    \centering
    \setlength{\tabcolsep}{2.5pt} 
    \begin{tabular}{cccccc}
         Original Image & 
        Original & 
         PGD $\ell_{2}$ & 
        PGD $\ell_{\infty}$ & 
        CFE $\ell_2$ & 
         p-CFE $\ell_0$ \\
    \includegraphics[width=0.14\textwidth]{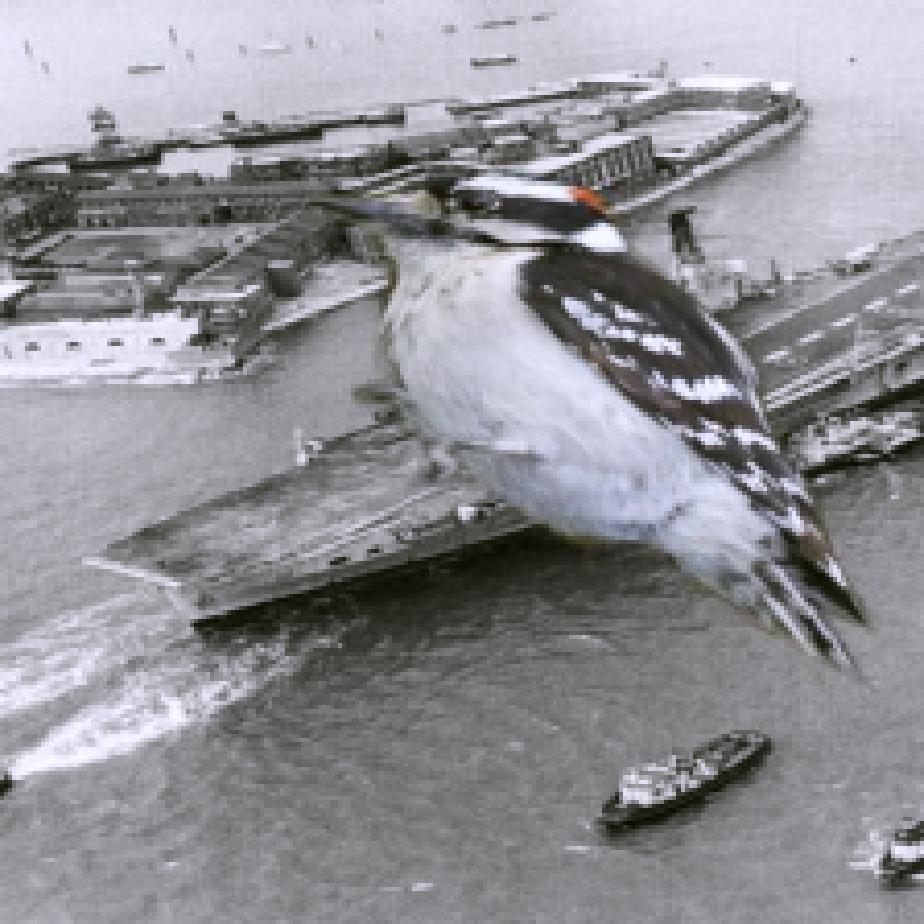} &
        \includegraphics[width=0.14\textwidth]{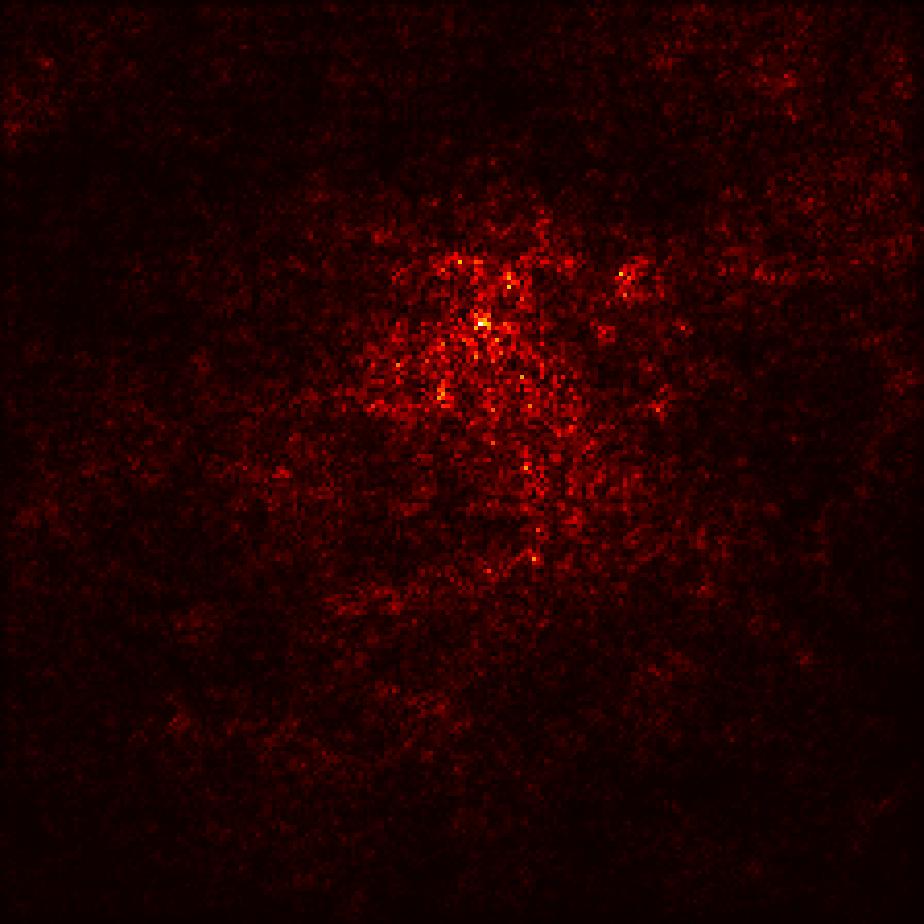} &
        \includegraphics[width=0.14\textwidth]{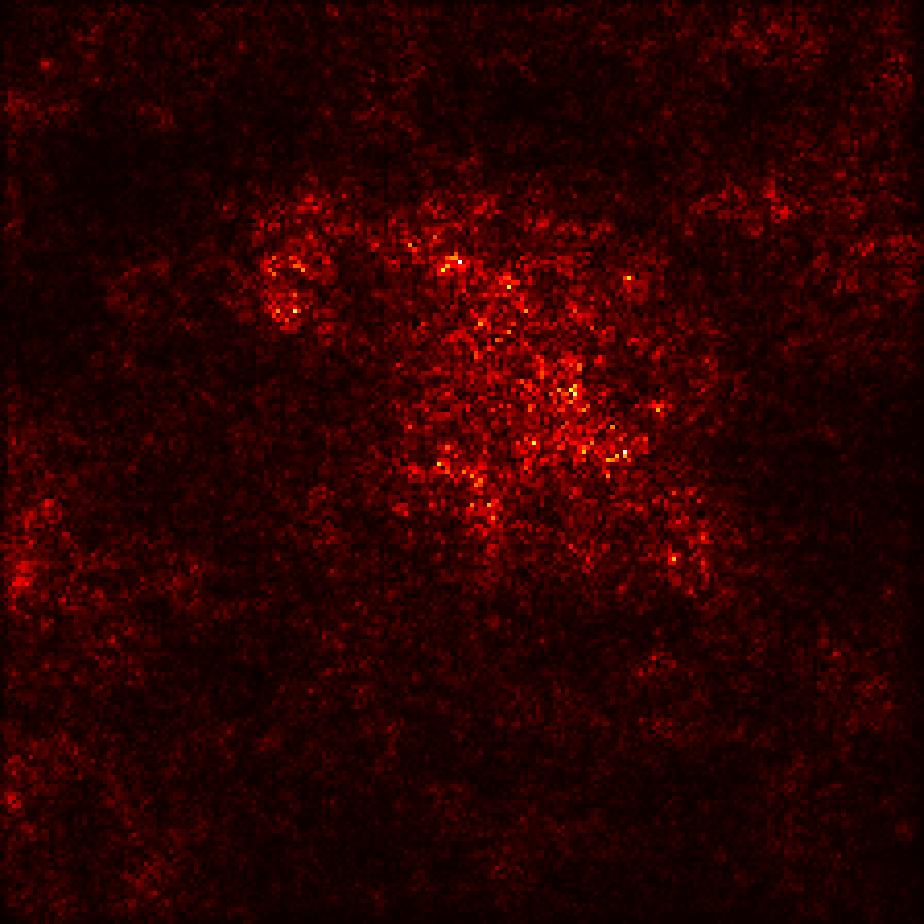} &
        \includegraphics[width=0.14\textwidth]{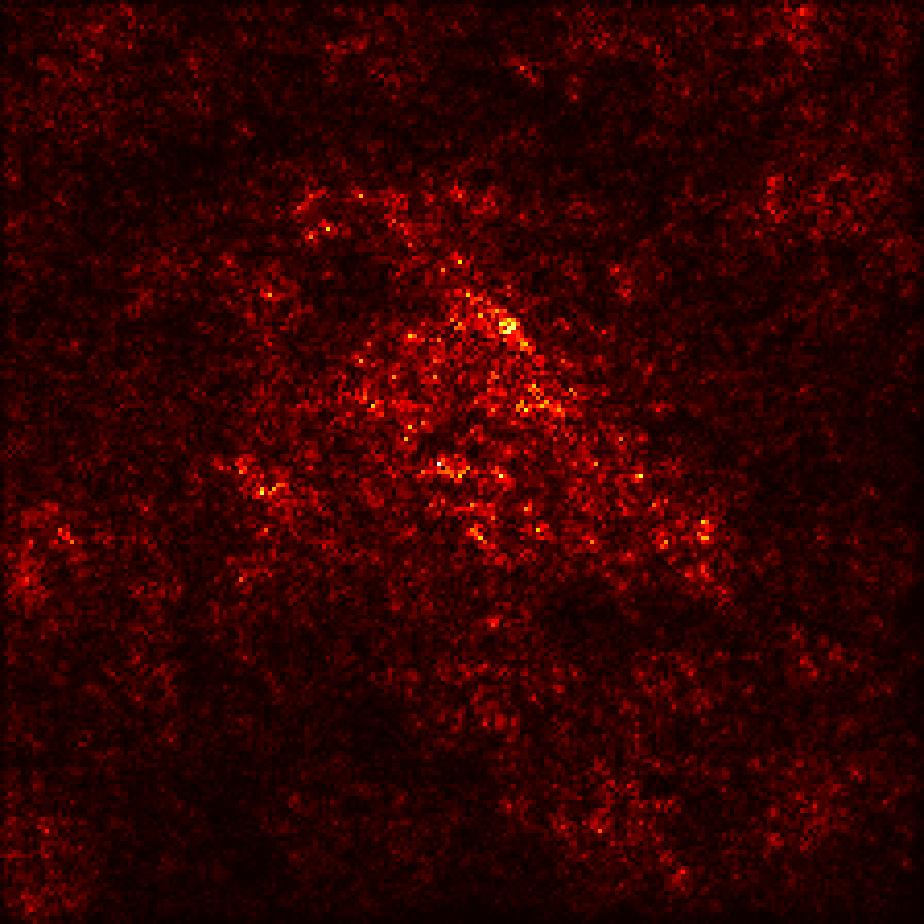} &
        \includegraphics[width=0.14\textwidth]{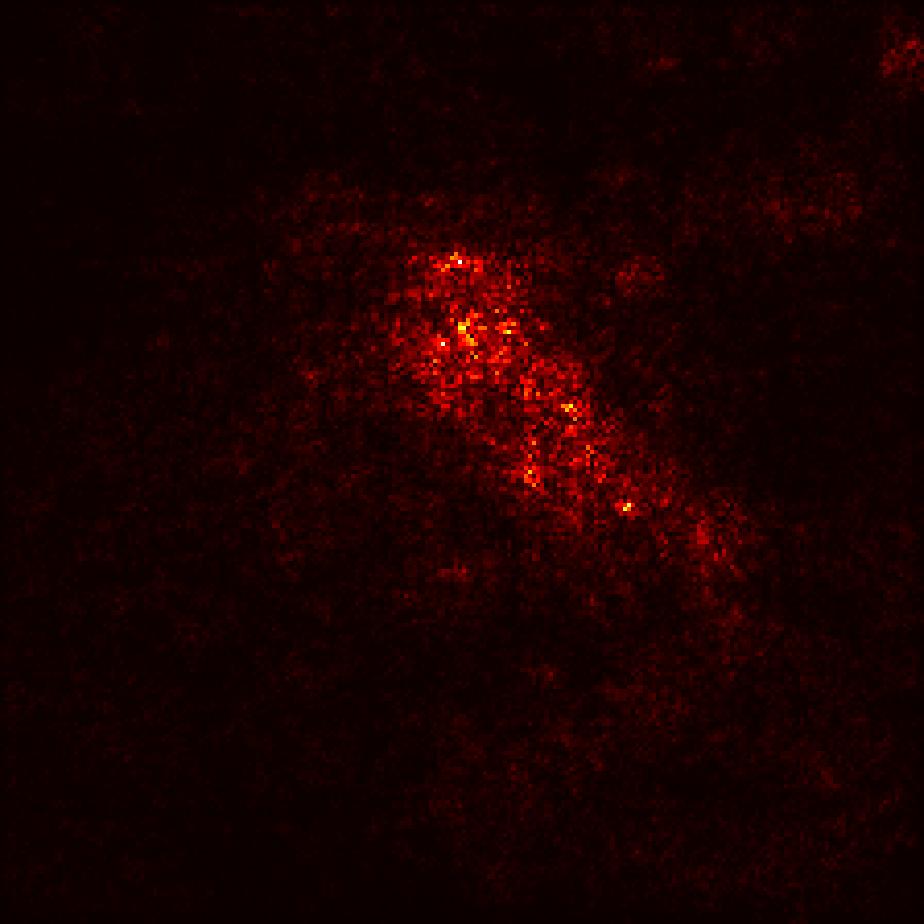} &
        \includegraphics[width=0.14\textwidth]{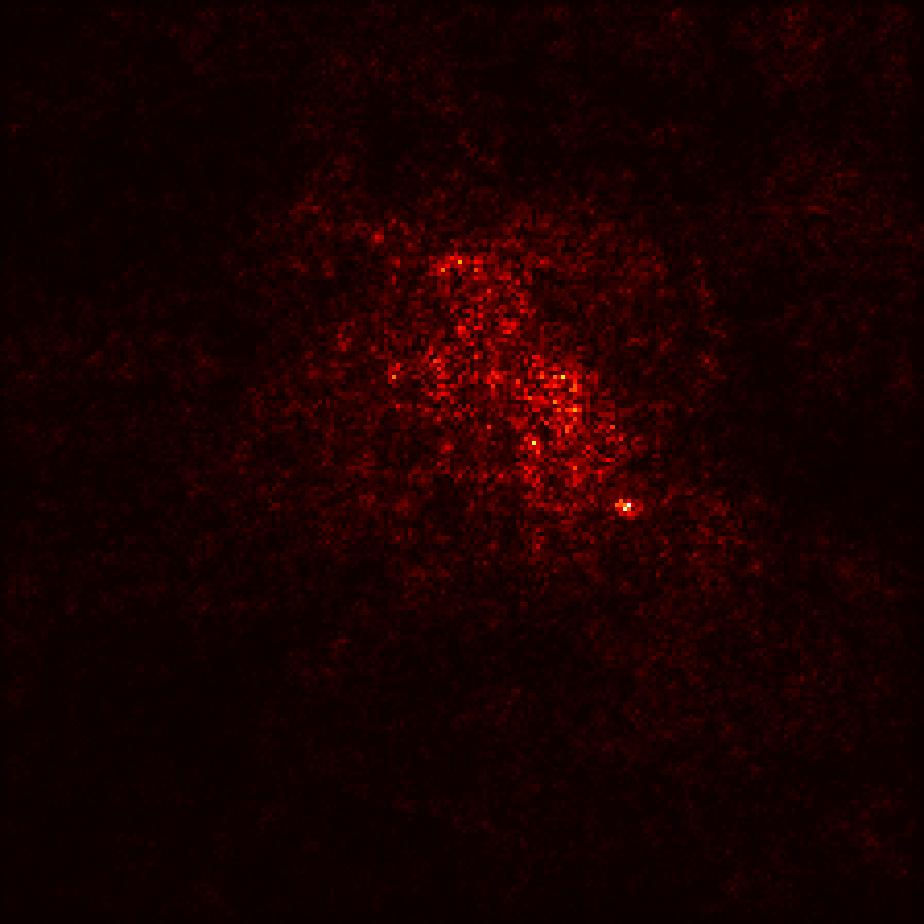} \\[2pt]
        \includegraphics[width=0.14\textwidth]{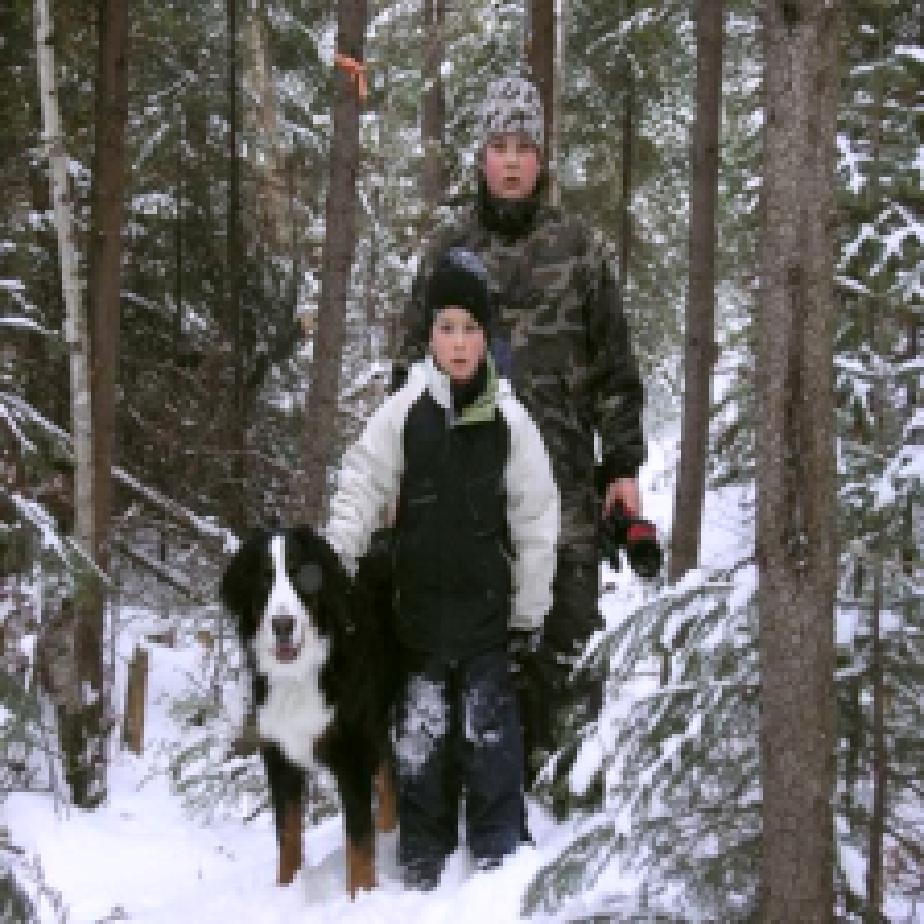} &
        \includegraphics[width=0.14\textwidth]{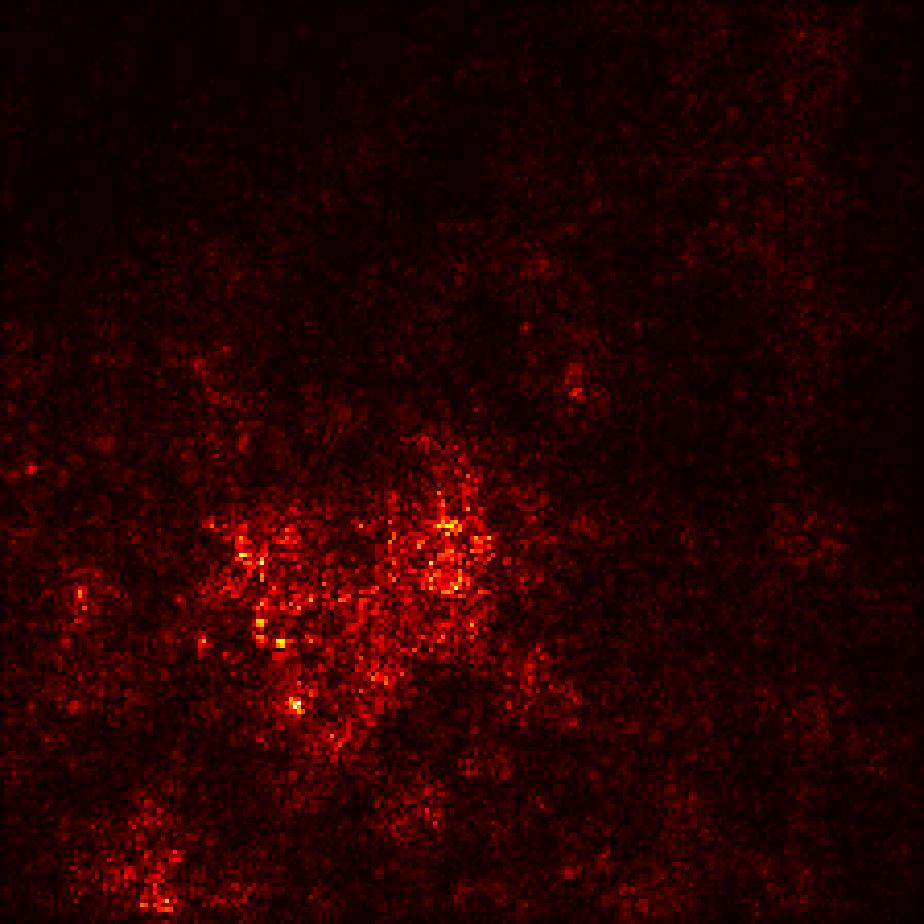} &
        \includegraphics[width=0.14\textwidth]{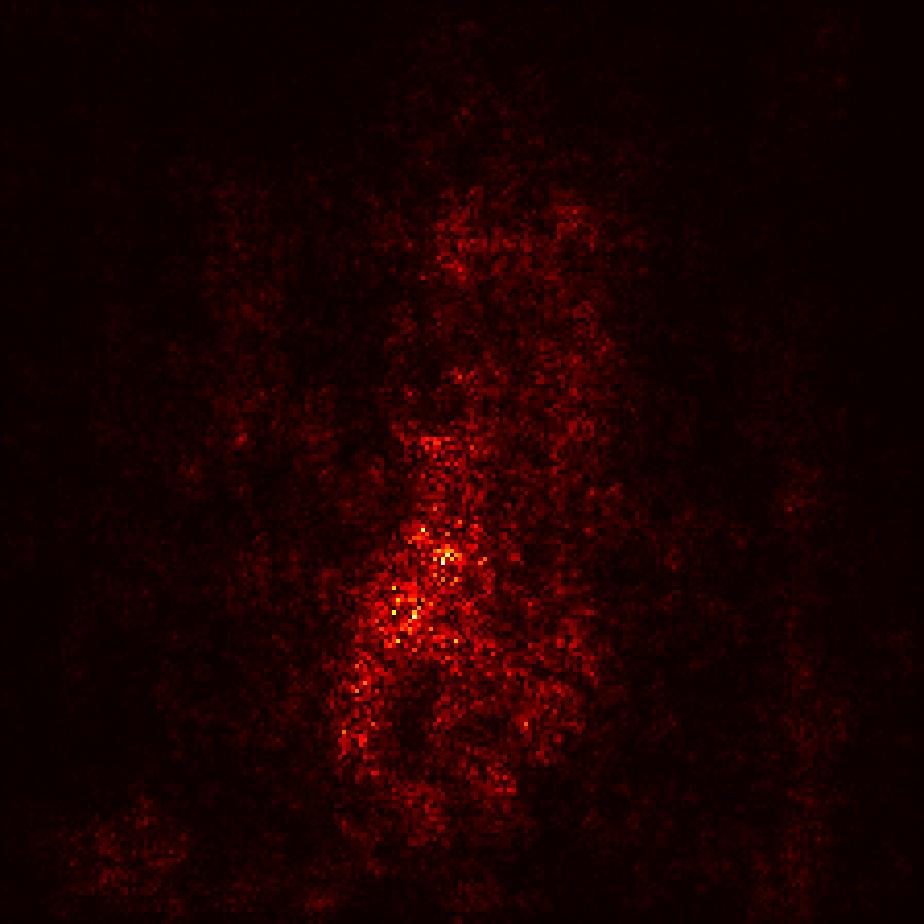} &
        \includegraphics[width=0.14\textwidth]{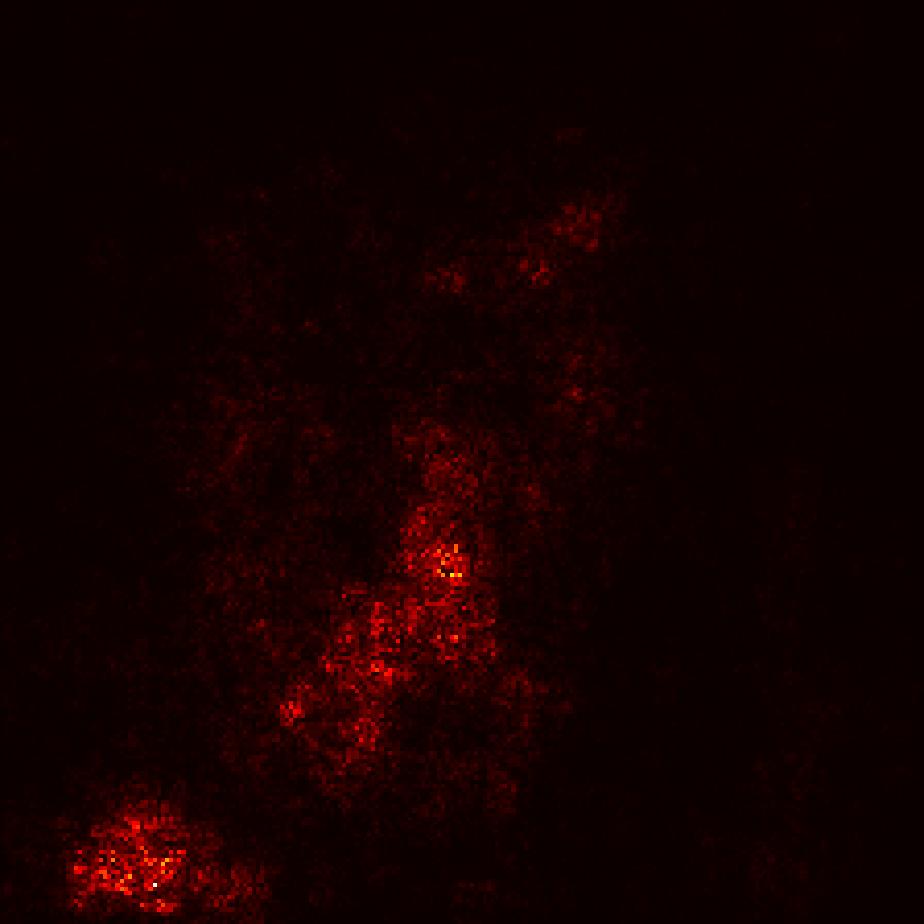} &
        \includegraphics[width=0.14\textwidth]{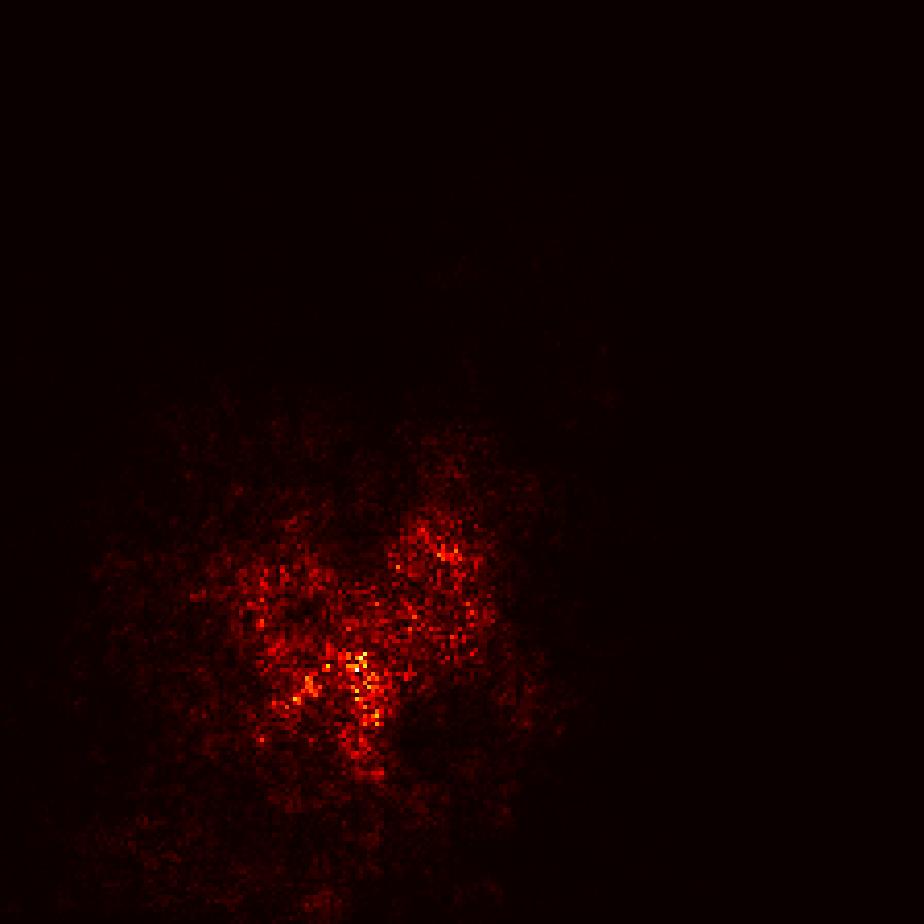} &
        \includegraphics[width=0.14\textwidth]{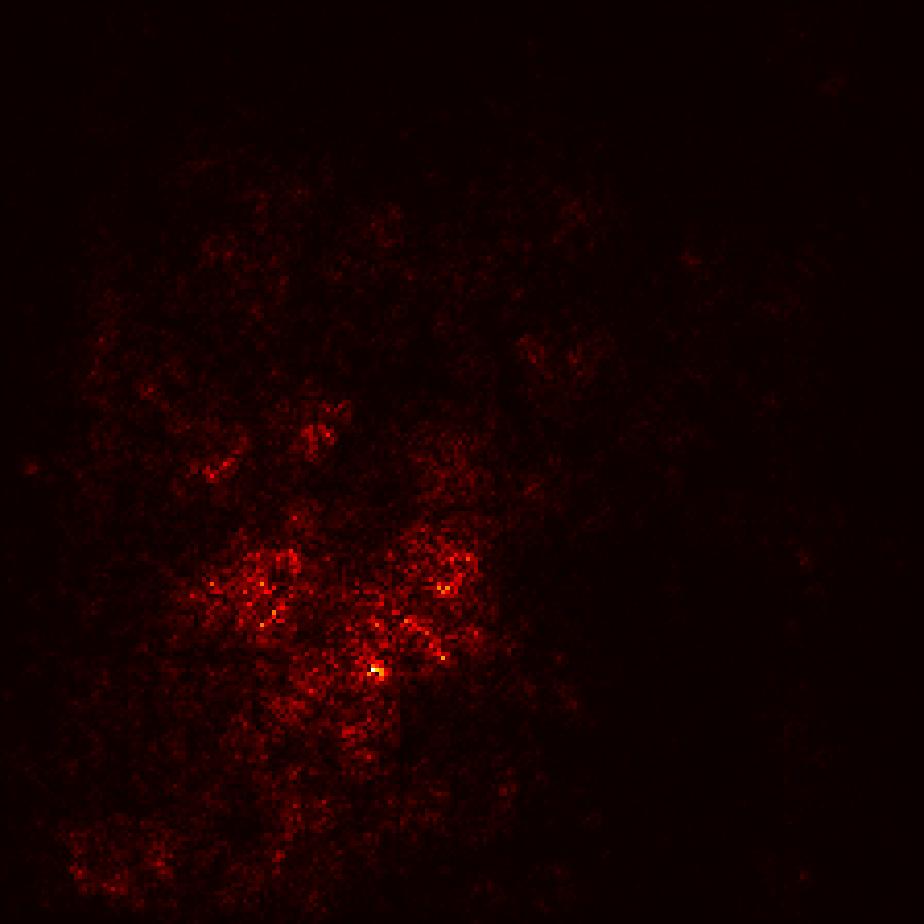}  
    \end{tabular} 
    
    \caption{Saliency maps for different models. Left to right: (1) original image of a land bird (top) and a dog (bottom), (2) saliency map from a standard model, (3-5) maps from models trained on PGD ($\ell_2$, $\ell_\infty$) and CFE $\ell_2$ adversarial examples, (6) maps from models trained on p-CFE $\ell_0$ examples. All models use ResNet50.}
    \label{fig:waterbirds_spucodogs_Grad-CAM}
\end{figure*}

\paragraph{Setup.} We fine-tune a pre-trained ResNet-50 model \citep{he2016deep} on original images, PGD ($\ell_2$, $\ell_\infty$), CFE $\ell_2$-adversarial perturbations~\citep{madry2018towards, wachter2017counterfactual}, and p-CFE ($\ell_0$) perturbations~\citep{2024_SadikuEtAl_Counterfactualexplanations}.\footnote{\citet{kumano2024theoretical} extended \cref{learningfromperturbations} to show that training solely on adversarial perturbations can match the accuracy of models trained on original data. Our experiments with this setup and standard vision datasets are detailed in \cref{additionallearningfromperturbations}.} The target classes $\tilde{y}$ are chosen uniformly at random, which makes the features of original images become uncorrelated with the labels. Training examples are shown in \cref{trainingdata}. We use the SGD optimizer in PyTorch with a learning rate of 1e-5, momentum of 0.9, weight decay of 5e-2, batch size of 8, and train for 360 epochs. 

\paragraph{Metrics.}
Train and test accuracies are the fractions of correctly classified samples in the training and test sets, respectively. To quantify spurious correlations, we use worst-group accuracy (WGA) as defined in \citep{sagawa2020investigation}. For instance, for WaterBirds, groups are defined as \emph{(attribute, label)} pairs. The worst-group—\emph{(waterbirds, land)}—has only 56 training samples, while other groups (e.g., \emph{(landbirds, land), (waterbirds, water), (landbird, water)} have up to 20 times more. A WGA of 0 indicates that all \emph{(waterbirds, land)} examples were misclassified, revealing a strong spurious correlation with the land background. We compute these metrics across five classifiers trained on original data, PGD ($\ell_2$, $\ell_\infty$), CFE ($\ell_2$), and p-CFE $\ell_0$ examples.

\begin{table*}[t]
    \caption{Train and test accuracies across various methods for various datasets. The model used is ResNet50 and we follow the setup of \citet{yang2023change}. The results of standard training (Original) are provided for reference. Learning from p-CFEs suffers less from spurious correlations than learning from adversarial perturbations. }
    \centering
    \small
    \begin{tabular}{cccccc|c}
    \toprule
        Dataset & Split Set & PGD $\ell_{2}$ (std) & PGD $\ell_{\infty}$ (std) & CFE $\ell_2$ (std) & p-CFE $\ell_0$ (std) & Original (std) \\
    \midrule
        \multirow{2}{*}{\begin{tabular}{c} WaterBirds  \end{tabular}} & Train & 97.04 (0.07) & \textbf{98.00} (0.08) & 97.93 (0.07) & 91.50 (0.56) & {99.98 (0.02)} \\
        & Test & 86.08 (0.69) & 86.02 (0.56) & \textbf{88.58} (0.61) & 86.54 (0.26) & { 87.56 (0.21)} \\
    \midrule
        \multirow{2}{*}{\begin{tabular}{c} SpuCoAnimals \end{tabular}} & Train & 96.25 (0.03) & 97.47 (0.02) & \textbf{97.89} (0.08) & 97.10 (0.01) & { 99.86 (0.18)} \\
        & Test & 78.10 (0.92) & 79.43 (0.68) & 79.00 (0.93) & \textbf{81.78} (0.59) & { 83.13} (0.37) \\
    \bottomrule
    \end{tabular}
    \label{tab:accuracies}
\end{table*}

\begin{table*}[!t]
    \caption{Worst-group accuracies across various methods for different datasets, following the configuration of \citet{yang2023change}. The results of standard training (Original) are provided for reference. Learning from p-CFEs suffers less from spurious correlations and even outperforms standard training on the WaterBirds dataset.}
    \centering
    \small
    \begin{tabular}{cccccc|c}
    \toprule
        Dataset & Split Set & PGD $\ell_{2}$ (std) & PGD $\ell_{\infty}$ (std) & CFE $\ell_2$ (std) & p-CFE $\ell_0$ (std) & Original (std) \\
    \midrule
        \multirow{2}{*}{\begin{tabular}{c} WaterBirds  \end{tabular}} & Train & 56.55 (4.20) & 72.00 (4.69) & 74.99 (2.54) & \textbf{77.97} (2.22) & 99.90 (0.13) \\
        & Test & 56.58 (2.17) & 61.72 (2.50) & 63.04 (2.19) & \textbf{76.05} (1.45) & 64.97 (1.45) \\
    \midrule
        \multirow{2}{*}{\begin{tabular}{c} SpuCoAnimals \end{tabular}} & Train & 62.60 (1.33) & 74.60 (1.41) & 72.86 (1.61) & \textbf{80.20} (0.86) & 99.70 (0.14) \\
        & Test & 56.06 (1.99) & 57.53 (1.79) & 56.60 (3.15) & \textbf{63.53} (1.55) & 65.60 (0.90) \\
    \bottomrule
    \end{tabular}
    \label{tab:worstgroupacc}
\end{table*}

\paragraph{Results.}From \cref{tab:accuracies}, learning from p-CFEs matches learning from perturbations (with $\ell_{2}$ and $\ell_{\infty}$ PGD) as well as CFE $\ell_2$ on achieving comparable training accuracy to standard training, thus extending empirical findings of \citet{kumano2024theoretical} to p-CFEs. 
Note that except for standard training, the models are trained on perturbed images with target incorrect labels~(cf.~\cref{learningfromperturbations}), and the training accuracy was evaluated on the original training images and labels. 
Moreover, \cref{tab:worstgroupacc} shows that on WaterBirds and SpuCoAnimals, training with p-CFEs substantially boosts worst-group accuracy—surpassing even standard (noise-free) training by 12\,\% on WaterBirds. This indicates that p-CFE-trained models rely less on spurious background features. \cref{fig:waterbirds_spucodogs_Grad-CAM} confirms that models trained on other perturbations focus heavily on spurious backgrounds, while p-CFE training shifts attention to relevant features (e.g., the bird and the dog), effectively mitigating spurious correlations.

\section{Conclusion and Discussion}
We showed that training with p-CFEs provides a compelling alternative to adversarial perturbations, guiding models toward semantically meaningful features and reducing reliance on spurious correlations. Our approach is data-efficient, model-agnostic, and requires no group labels. Future work includes scaling to higher-dimensional datasets using more expressive density estimators, and extending to large models such as LLMs and vision-language models (VLMs).
Further, combining the theoretical framework of learning from adversarial perturbations in \citep{kumano2024theoretical,kumano2024wide} with the connection between adversarial perturbations and p-CFEs~\citep{pawelczyk2022exploring,freiesleben2022intriguing} can justify our observations.

\section*{Acknowledgement}
This research was partially supported by the DFG Cluster of
Excellence MATH+ (EXC-2046/1, project id 390685689)
funded by the Deutsche Forschungsgemeinschaft (DFG) as
well as by the German Federal Ministry of Education and
Research (fund number 01IS23025B). Hiroshi Kera was supported by JSPS KAKENHI Grant Number JP23KK0208.

\bibliographystyle{plainnat}
\bibliography{references}

\clearpage
\appendix
\onecolumn
\section*{Appendix}
\addcontentsline{toc}{section}{Appendices}
\renewcommand{\thesubsection}{\Alph{subsection}}
\section{Spurious Correlations - Additional Examples}
\label{spuriousadditional}
\cref{fig:waterbirdsspuco_Grad-CAM_ext} illustrates additional examples of spurious correlations. In the top row, the model relies on the presence of water to classify waterbirds. In the bottom row, it associates the outdoor background with the presence of big dogs.
\def\offs{1.9}
\def\imgwidth{0.165}
\begin{figure}
    \centering
    \scalebox{1.2}{
    \begin{tikzpicture}

        \node at (\offs, \offs + 1.2) {\small Original};
        \node at (2.55*\offs, \offs + 1.235) {\small Grad-CAM};
    
        \node at (\offs, \offs) {\includegraphics[width=\imgwidth\textwidth, height=0.10\textwidth]{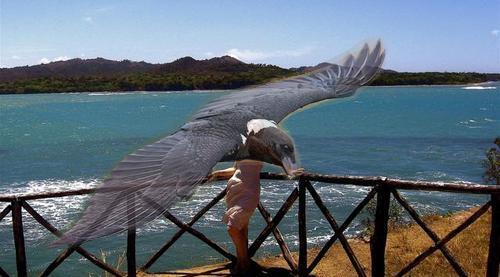}};
        \node at (2.55*\offs, \offs) {\includegraphics[width=\imgwidth\textwidth, height=0.10\textwidth]{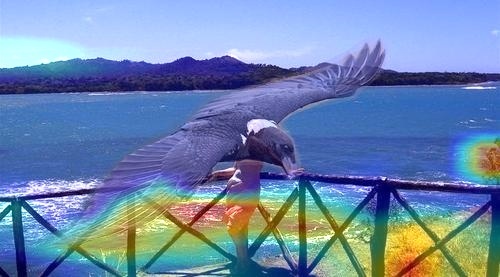}};

        \node at (\offs, 0.016*\offs) {\includegraphics[width=\imgwidth\textwidth, height=0.10\textwidth]{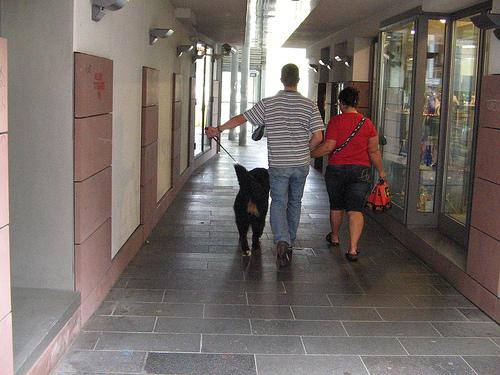}};
        \node at (2.55*\offs, 0.016*\offs) {\includegraphics[width=\imgwidth\textwidth, height=0.10\textwidth]{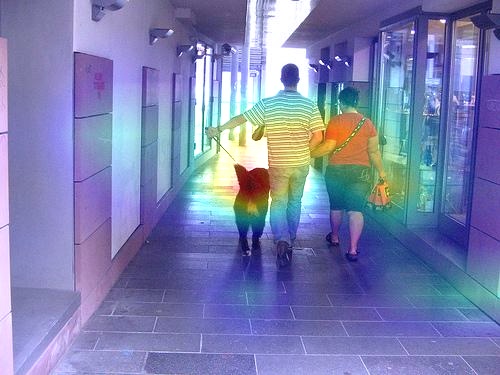}};
    \end{tikzpicture}   
    }
    \caption{\textbf{Top row}: Original and Grad-CAM visualizations for a misclassified \emph{landbird} (with a water background) from the WaterBirds dataset—incorrectly predicted as a \emph{waterbird}. \textbf{Bottom row}: Original and Grad-CAM visualizations for a misclassified \emph{small dog} (with an outdoor background) from the SpuCoAnimals dataset—incorrectly predicted as a \emph{big dog}.}
    \label{fig:waterbirdsspuco_Grad-CAM_ext}
\end{figure}

\section{Learning from p-CFEs - Additional Experiments}
\label{additionallearningfromperturbations}
In this section, we extend the work of \citet{kumano2024theoretical} on learning from adversarial perturbations to traditional $\ell_2$ CFEs from \citet{wachter2017counterfactual}, as well as the more recent $\ell_0$ p-CFEs proposed by \citet{2024_SadikuEtAl_Counterfactualexplanations}. We generate adversarial examples using Projected Gradient Descent (PGD)  \citep{madry2018towards} with cross-entropy loss under varying norms ($\ell_{2}, \ell_{\infty}$).  CFEs are constructed by minimizing the cross-entropy loss regularized by the unweighted squared Euclidean distance, controlled by the tradeoff parameter $\lambda$. We denote $\lambda_{CF}$ as the learning rate used by the Adam optimizer during CFE optimization. For $\text{p-CFE} \ \ell_{0}$, we define $L$ as the Lipschitz constant, and $\lambda_{steps}$ as the number of search steps.

\subsection{Results on Artificial Data}

\begin{figure}[H]
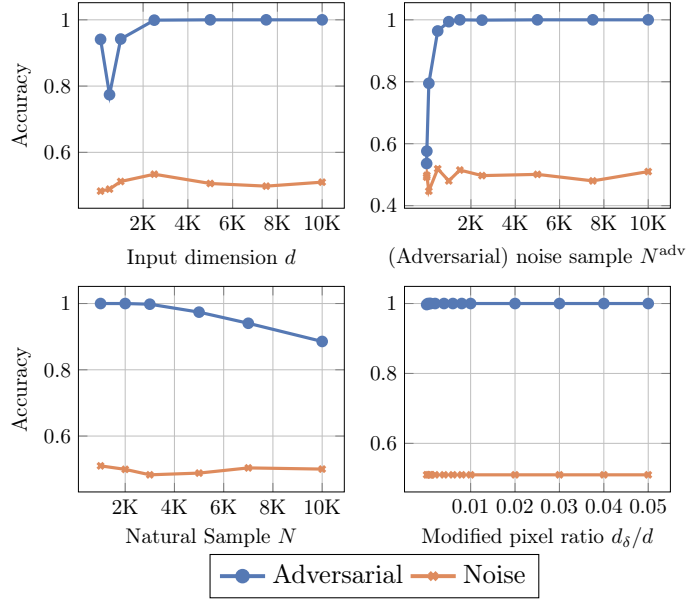

    \centering

    
    \normalsize
    \caption{Comparison of the accuracy of the model trained on CFE $\ell_2$ perturbations and noise trained model on the clean dataset. Data was acquired from a \textbf{uniform} distribution. The hyper-paramaters $\lambda=0.001$ and $\lambda_{CF} = 0.01$ were used. The algorithm was run for $0.05d$ iterates, where $d$ is the input dimension.}
    \label{uniformcfe}
    \end{figure}

\begin{figure}[H]
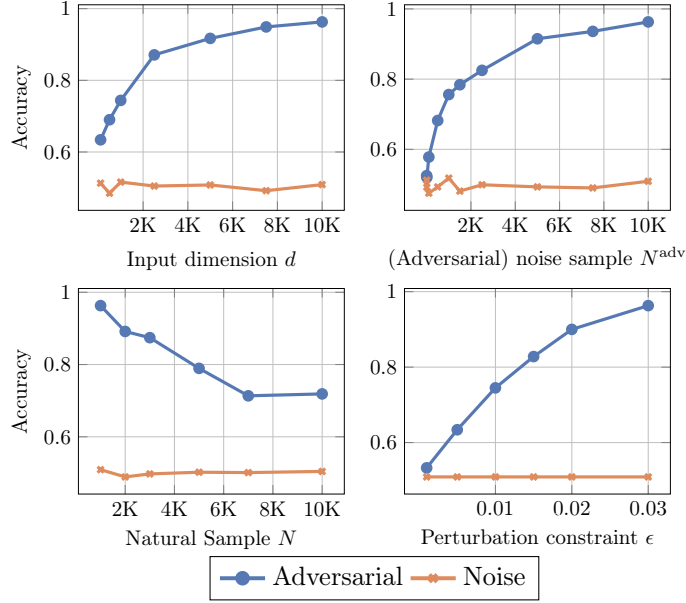

    \centering
    %
    
    \normalsize
    \caption{Comparison of the accuracy of the model trained on PGD $\ell_\infty$ perturbations and noise trained model on the clean dataset. Data was acquired from a \textbf{Gaussian} distribution. The hyper-parameter $\epsilon=0.03$ was used, along with 100 iterations. The hyper-parameter $\epsilon=0.03$ was used for most of the graphs (top-left, top-right, bottom-left). For the graph at bottom-right, multiple perturbation constraints were used.}
    \label{gaussianpgdinf}
    \end{figure}

We experiment on 2D dim data generated from uniform or Gaussian distribution. The model used for these experiments is a one-layer neural network \citep{kumano2024theoretical}.
\crefrange{uniformcfe}{gaussianpgdinf} compare the accuracies of models trained with adversarial examples versus noise-augmented data, across varying input dimensions and numbers of natural and adversarial samples. For CFE $\ell_2$ and p-CFE $\ell_{0}$, we additionally vary the ratio of modified pixels, denoted by $d_{\delta}/d$ where $d_{\delta}$ is the number of modified pixels and $d$ is the total number of pixels. These results extend the findings of \citet{kumano2024theoretical} to settings involving both traditional CFEs and p-CFEs.

    \begin{figure}[H]
    \centering
    %
    
    \normalsize
    \caption{Comparison of the accuracy of the model trained on CFE $\ell_2$ perturbations and noise trained model on the clean dataset. Data was acquired from a \textbf{Gaussian} distribution. The hyper-paramaters $\lambda=0.001$ and $\lambda_{CF} = 0.01$ were used. The algorithm was run for $0.05d$ iterates, where $d$ is the input dimension.}
    \end{figure}

    \begin{figure}[H]
    \centering
    %
    
    \normalsize
    \caption{Comparison of the accuracy of the model trained on p-CFE $\ell_0$ perturbations and noise trained model on the clean dataset. Data was acquired from a \textbf{uniform} distribution. The hyper-parameters $\lambda_{steps}=5, L=1.0 \ \text{and} \ 0.05d$ iterations were considered. Here, $\lambda_{steps} \ \text{and} \ L$ denote the number of search steps and step size, respectively.}
    \end{figure}

    \begin{figure}[H]
    \centering
    %
    
    \normalsize
    \caption{Comparison of the accuracy of the model trained on p-CFE $\ell_0$ perturbations and noise trained model on the clean dataset. Data was acquired from a \textbf{Gaussian} distribution. The hyper-parameters $\lambda_{steps}=5, L=1.0 \ \text{and} \ 0.05d$ iterations were considered. Here, $\lambda_{steps} \ \text{and} \ L$ denote the number of search steps and step size, respectively.}
    \end{figure}

    \begin{figure}[H]
    \centering
    %
    
    \normalsize
    \caption{Comparison of the accuracy of the model trained on PGD $\ell_2$ perturbations and noise trained model on the clean dataset. Data was acquired from a \textbf{uniform} distribution. The hyper-parameter $\epsilon=0.78$ was used for most of the graphs (top-left, top-right, bottom-left). For the graph at bottom-right, multiple perturbation constraints, along with multiple input dimensions, were used. All the attacks were executed for 100 iterations.}
    \end{figure}

    \begin{figure}[H]
    \centering
    %
    
    \normalsize
    \caption{Comparison of the accuracy of the model trained on PGD $\ell_2$ perturbations and noise trained model on the clean dataset. Data was acquired from a \textbf{Gaussian} distribution. The hyper-parameter $\epsilon=0.78$ was used for most of the graphs (top-left, top-right, bottom-left). For the graph at bottom-right, multiple perturbation constraints, along with multiple input dimensions, were used. All the attacks were executed for 100 iterations.}
    \end{figure}
    
    \begin{figure}[H]
    \centering
    %
    
    \normalsize
    \caption{Comparison of the accuracy of the model trained on PGD $\ell_\infty$ perturbations and noise trained model on the clean dataset. Data was acquired from a \textbf{uniform} distribution. The hyper-parameter $\epsilon=0.03$ was used, along with 100 iterations. The hyper-parameter $\epsilon=0.03$ was used for most of the graphs (top-left, top-right, bottom-left). For the graph at bottom-right, multiple perturbation constraints were used.}
    \end{figure}

\subsection{Results on Original (Natural) Datasets}

A convolutional neural network was used for the MNIST \citep{deng2012mnist} and Fashion-MNIST (FMNIST) \citep{xiao2017fashion} datasets, while a WideResNet was used for CIFAR-10 \citep{krizhevsky2009learning}, following the setup of \citet{kumano2024theoretical}. We denote deterministic target labels by D and random target labels by R. The learning rate for the Adam optimizer used in CFE generation is set to $\lambda_{CF} = 0.01$, unless stated otherwise. All algorithms were run for 100 iterations by default. 

\begin{figure}[H]
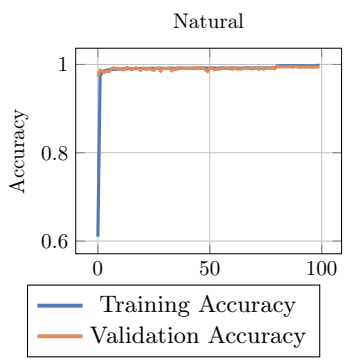

    \centering


    \centering
    {\small \pgfplotslegendfromname{legend_pltmnist}}

    \caption{Standard training and validation accuracy of a simple convolutional neural network model trained on MNIST dataset for 100 epochs.}
    \label{mniststandard}
\end{figure}

For each dataset, we first train a model on the clean training set, then use it to generate adversarial samples or CFEs from that same set. A second model is then trained on the perturbed data. \crefrange{mniststandard}{cifarpgdinf} compare the training and validation accuracies of models trained on clean versus perturbed data. Validation accuracy refers to performance on the clean validation set, while training accuracy reflects performance on the perturbed training data. These results extend the findings of \cite{kumano2024theoretical} to settings involving both traditional CFEs and p-CFEs for standard benchmark datasets.

\subsubsection{MNIST}
    
    \begin{figure}[H]
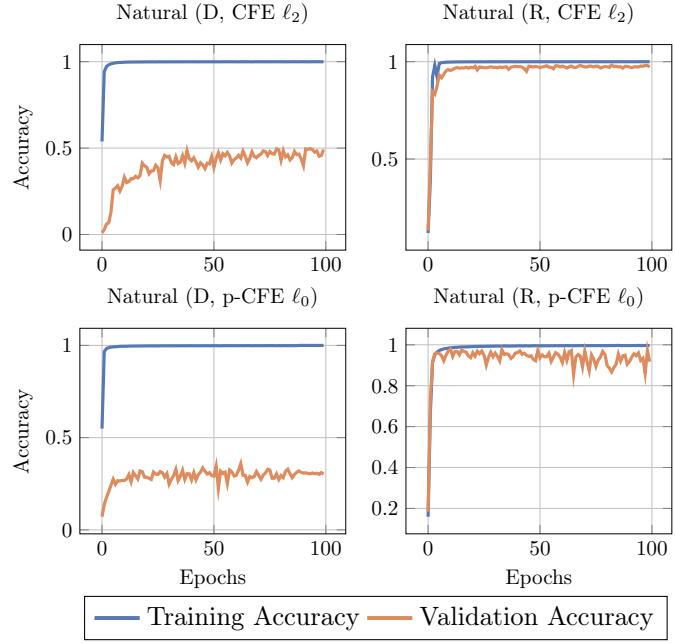

    \centering
    %
    \normalsize
    \caption{Training and Validation accuracies for models trained on MNIST counterfactuals generated using p-CFE $\ell_{0}$ and CFE $\ell_{2}$. For CFE $\ell_{2}$, we used $\lambda=0.1$ ($\ell_2$ (regularization strength)) and $\lambda_{CF} = 0.01$ (step-size for Adam Optimizer). For p-CFE $\ell_{0}$, we used a Lipschitz constant $L=$1e-4 and 5 search steps. Both methods were run for 100 epochs.}
    \end{figure}
    
    \begin{figure}[H]
    \centering
    %
    \normalsize
    \caption{Training and Validation accuracies for models trained on MNIST adversarial samples generated by PGD $\ell_{\infty}$ and $\ell_{2}$. The hyper-parameters $\epsilon=2 \ \text{and} \ \epsilon=0.3$ were used for the PGD $\ell_{2}$ and $\ell_{\infty}$ attacks, respectively. Both methods were run for 100 epochs.}

    \end{figure}

\subsubsection{FMNIST}
    \begin{figure}[H]
    \centering
    %

    \centering
    {\small \pgfplotslegendfromname{legend_pltfmnist}}

    \caption{Standard training and validation accuracy of a simple convolutional neural network model trained on FMNIST dataset for 200 epochs.}
\end{figure}

    \begin{figure}[H]
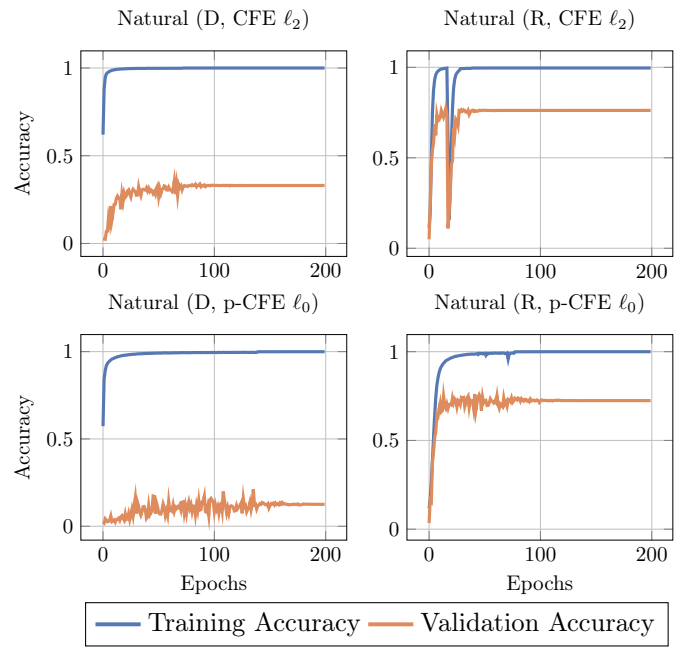

    \centering
    %
    \normalsize
    \caption{Training and Validation accuracies for models trained on FMNIST adversarial samples generated using CFE $\ell_{2}$ and p-CFE $\ell_{0}$. For CFE we used $\ell_{2}$ $\lambda=0.01 \ \text{and} \ \lambda_{CF} = 0.01$; for p-CFE $\ell_{0}$, a Lipschitz constant $L=$1e-5 and 5 search steps were considered. Both methods were run for 100 epochs.}

    \end{figure}

    \begin{figure}[H]
    \centering
    %
    \normalsize
    \caption{Training and Validation accuracies for models trained on FMNIST adversarial samples generated by PGD $\ell_{\infty}$ and $\ell_{2}$. The hyper-parameters $\epsilon=2 \ \text{and} \ \epsilon=0.3$ were considered for PGD $\ell_{2}$ and $\ell_{\infty}$ attack, respectively. Both methods were run for 100 epochs.}

    \end{figure}

\subsubsection{CIFAR-10}
    \begin{figure}[H]
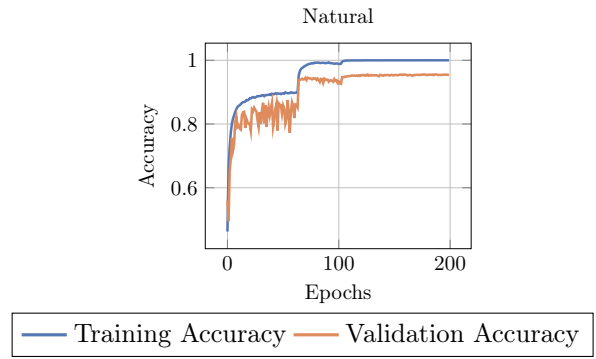

    \centering
    %

    \centering
    {\pgfplotslegendfromname{legend_pltcifar10}}
    \normalsize
    \caption{Standard training and validation accuracy of a wide ResNet model trained on the CIFAR10 dataset for 200 epochs.}
\end{figure}

    \begin{figure}[H]
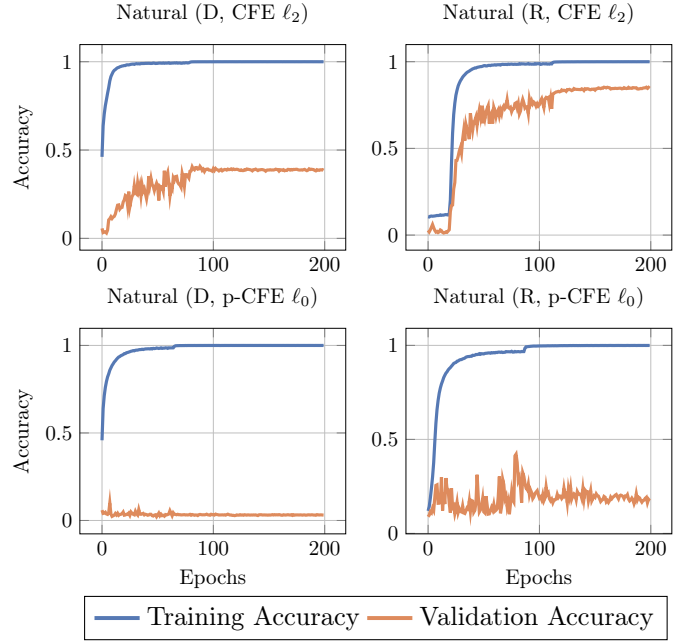

    \centering
    %
    \normalsize
    \caption{Training and Validation accuracies for models trained on CIFAR10 counterfactuals generated by CFE $\ell_{2}$ and p-CFE $\ell_{0}$. The hyper-parameters $\lambda=0.01 \ \text{and} \ \lambda_{CF} = 0.001$ were considered for CFE $\ell_{2}$, and for p-CFE $\ell_{0}$, Lipschitz constant $L=$1e-4 with 4 search steps were considered. Both methods were run for 100 epochs.}

    \end{figure}

    \begin{figure}[H]
    \centering
    %
    \normalsize
    \caption{Training and Validation accuracies for models trained on CIFAR10 adversarial samples generated by PGD $\ell_{\infty}$ and $\ell_{2}$. The hyper-parameters $\epsilon=0.5 \ \text{and} \ \epsilon=0.1$ were considered for PGD $\ell_{2}$ and $\ell_{\infty}$ attack, respectively. Both methods were run for 100 epochs.}
    \label{cifarpgdinf}
    \end{figure}

\end{document}